\documentclass[journal]{IEEEtran}  

\usepackage{booktabs}  
\usepackage{threeparttable}  
\usepackage{array}
\usepackage{multicol}  
\usepackage{multirow}
\usepackage{makeidx}  
\usepackage[pdftex,bookmarksnumbered, colorlinks, linkcolor=red,anchorcolor=green,citecolor=green]{hyperref} 
\usepackage[pdftex]{graphicx} 
\usepackage{caption}
\usepackage{subfigure}
\usepackage{url}
\usepackage{amsmath}
\usepackage{amsfonts}
\usepackage{graphicx} 
\usepackage{adjustbox}
\usepackage[usenames,dvipsnames]{pstricks}
\usepackage{epsfig}
\usepackage{pst-grad} 
\usepackage{pst-plot} 
\usepackage{bbding}
\usepackage{color, colortbl}
\usepackage{diagbox} 
\usepackage{algorithm}
\usepackage{algpseudocode}
\usepackage{textcomp}
\usepackage{tikz}
\usepackage{cite}

\newcommand{\rev}[1]{\textcolor{black}{#1}}

\title{ \LARGE \bf
		Pose Estimation for Ground Robots: On Manifold Representation, Integration, Re-Parameterization, and Optimization}
\author{Mingming Zhang$^{1\dag}$, Xingxing Zuo$^{1,2\dag}$, Yiming Chen$^1$,  Yong Liu$^2$ and Mingyang Li$^{1\ddag}$
\\ 
$^1$Alibaba Group \\
{\tt \{ mingmingzhang, xingzuo,
	yimingchen, mingyangli \}@alibaba-inc.com } \\
$^2$Institute of Cyber-System and Control, Zhejiang University, China \\
\tt {yongliu}@iipc.zju.edu.cn
\thanks{$\dag$Mingming Zhang and Xingxing Zuo are joint first authors with equal contribution to this work.
}
\thanks{$\ddag$Mingyang Li is the corresponding author.}
}

\begin{document}

\maketitle

\begin{abstract}
In this paper, we focus on motion estimation dedicated for \rev{non-holonomic} ground robots, 
by probabilistically fusing measurements from the wheel odometer and \rev{exteroceptive sensors}.
For ground robots, the wheel odometer is widely used in pose estimation tasks, especially in applications under planar-scene based environments. However, 
since the wheel odometer only provides 2D motion estimates, it is extremely challenging to use that for performing accurate full 6D pose (3D position and 3D \rev{orientation}) estimation.
Traditional methods on 6D pose estimation either approximate sensor or motion models, 
at the cost of accuracy reduction, or rely on other sensors, e.g., inertial measurement unit (IMU), \rev{to provide complementary  measurements}. 
By contrast, in this paper, \rev{we propose a 
novel method to utilize the wheel odometer for 6D pose estimation, by modeling and utilizing motion manifold for ground robots. Our approach is probabilistically formulated and only requires the wheel odometer and  an exteroceptive sensor (e.g., a camera).}
Specifically, our method 
i) formulates the motion manifold of ground robots by parametric representation, 
ii) performs manifold based 6D integration with the wheel odometer measurements only, and 
iii) re-parameterizes manifold equations periodically for error reduction.
To demonstrate the effectiveness and applicability of the proposed algorithmic modules, we integrate that into a sliding-window pose estimator \rev{ by using measurements from the wheel odometer and a monocular camera}.
By conducting extensive simulated and real-world experiments, we show that
the proposed algorithm outperforms \rev{competing}
state-of-the-art algorithms by a significant margin \rev{in pose estimation accuracy}, 
especially when deployed in \rev{complex} large-scale real-world environments.
\end{abstract}

\IEEEpeerreviewmaketitle

\newcommand{\bnn}{{\mathbf n}}
\newcommand{\bst}{{ \boldsymbol x}}
\newcommand{\bzeta}{{ \boldsymbol \zeta}}
\newcommand{\bsth}{{ \hat{\boldsymbol x}}}
\newcommand{\bstt}{ \tilde{\boldsymbol x}}
\newcommand{\bstd}{{ \dot{\boldsymbol x}}}
\newcommand{\bstdt}{{ \dot{\tilde{\boldsymbol x}}}}
\newcommand{\bO}{\mathbf{O}}
\newcommand{\bL}{\mathbf{L}}
\newcommand{\bT}{\mathbf{T}}
\newcommand{\bp}{\mathbf{p}}
\newcommand{\bl}{\mathbf{l}}
\newcommand{\ba}{\mathbf{a}}
\newcommand{\bomega}{\boldsymbol{\omega}}
\newcommand{\btheta}{\boldsymbol{\theta}}
\newcommand{\bo}{\mathbf{o}}
\newcommand{\bc}{\mathbf{c}}
\newcommand{\bz}{\mathbf{z}}
\newcommand{\bt}{\mathbf{t}}
\newcommand{\bQ}{\mathbf{Q}}
\newcommand{\bC}{\mathbf{C}}
\newcommand{\bA}{\mathbf{A}}
\newcommand{\bg}{\mathbf{g}}
\newcommand{\bB}{\mathbf{B}}
\newcommand{\bb}{\mathbf{b}}
\newcommand{\bG}{\mathbf{G}}
\newcommand{\bn}{\mathbf{n}}
\newcommand{\bI}{\mathbf{I}}
\newcommand{\bR}{\mathbf{R}}
\newcommand{\dbR}{\delta \mathbf{R}}
\newcommand{\dr}{\delta r}
\newcommand{\bD}{\mathbf{D}}
\newcommand{\bv}{\mathbf{v}}
\newcommand{\bJ}{\mathbf{J}}
\newcommand{\be}{\mathbf{e}}
\newcommand{\bx}{\mathbf{x}}
\newcommand{\dbp}{\delta \mathbf{p}}
\newcommand{\bm}{\mathbf{m}}
\newcommand{\bq}{\mathbf{q}}
\newcommand{\bphi}{\boldsymbol{\phi}}
\newcommand{\bE}{\mathbf{E}}

\section{Introduction}
The pose (position and \rev{orientation}) estimation problem for ground robots has been under active research and development for a couple of decades~\cite{yap2011particle,zhang2017low,levinson2007map,wolcott2014visual,hitcm2019sdf,censi2013simultaneous,cadena2016past,Zuo2019IROS}.
The most mature technique of computing poses for ground robots in large-scale environments is the one that relies on high-quality \rev{global positioning and inertial navigation systems (GPS-INS)} together with 3D laser range-finders~\cite{levinson2007map,wan2018robust,klingner2013street,levinson2011towards}. 
This design is widely used in autonomous driving vehicles to provide precise pose estimates for scene understanding, path planning, and decision \rev{making}~\cite{levinson2011towards}.
However, those systems are at high manufacturing and maintenance costs, requiring thousands or even tens or hundreds of thousands of dollars, which inevitably prevent their wide applications. Alternatively, low-cost pose estimation approaches have gained increased interests in recent years, especially the ones that rely on cameras~\cite{wolcott2014visual,mur2015orb}. Camera's size, low cost, and 3D sensing capability make itself a popular sensor. \rev{Another widely used sensor is IMU, which provides high-frequency estimates on rotational velocity and specific force of a moving platform.}
Since IMU and camera sensors have complementary characteristics, 
when used together with an IMU, the accuracy and robustness of vision-based pose estimation can be significantly improved~\cite{li2013high,qin2017vins,lynen2015get,li2013optimization,huai2018robocentric,liu2018ice,nisar2019vimo}. In fact, camera-IMU pose estimation\footnote{Camera-IMU pose estimation can also be termed as \rev{visual-inertial} localization~\cite{hesch2014camera}, vision-aided inertial navigation~\cite{mourikis2007}, visual-inertial odometry (VIO)~\cite{li2013high}, or visual-inertial navigation system (VINS)~\cite{qin2017vins} in other papers.} is widely used in real commercial applications, e.g., smart drones, augmented and virtual reality headsets, or mobile phones. 

However, all methods mentioned above are {\em not} optimized for ground robots.
Although \rev{visual-inertial pose estimation} generally performs better than camera only \rev{algorithms}
by resolving the ambiguities in estimating scale, roll, and pitch~\cite{li2013high,lynen2015get,li2014online},
it has its own limitations when used for ground robots.
Firstly, there are a couple of degenerate cases  that  can  result  in  large errors when performing motion estimation,  e.g., \rev{static motion}, 
zero rotational velocity motion,
constant local linear acceleration motion, and so on ~\cite{li2013high,wu2017vins,kottas2013detecting}.
The likelihood of encountering those degenerate cases on ground robots is significantly larger than that on hand-held mobile devices.
Secondly, unlike drones or smart headset devices that move freely in 3D space, ground robots can only move on a manifold (e.g., ground surfaces) due to the nature of their mechanical design.
This makes it possible to use additional low-cost sensors and derive extra mathematical constraints for improving the estimation performance~\cite{censi2013simultaneous,wu2017vins}.

When looking into the literature and applications of ground robot pose estimation, wheel odometer
is a widely used sensor system, for providing 2D linear and rotational velocities. Compared to IMUs, wheel odometer has two major advantageous factors in pose estimation. On the one hand, wheel odometer provides linear velocity directly, while IMU measures gravity affected linear accelerations. Integrating IMU measurements to obtain velocity estimates will inevitably suffer from measurement errors, integration errors, and state estimation errors (especially about roll and pitch). On the other hand, errors in positional and rotational estimates by integrating IMU measurements are typically a function of time. 
\rev{Therefore, integrating IMU measurements for a long-time period will inevitably lead to unreliable pose estimates, regardless of the robot's motion. 
However, when a robot moves slowly or keeps static, long-time pose integration is not a significant problem} by using the wheel odometer system, due to its nature of generating measurements by counting the wheel rotating impulse.

However, the majority of existing work that uses wheel odometer for pose estimation focus on `planar surface' applications, which are typically true {\em only} for indoor environments~\cite{censi2013simultaneous,wu2017vins,quan2018tightly}. 
While there are a couple of approaches for performing 6D pose estimation using wheel odometer measurements, the information utilization in those methods \rev{appears} to be {\em sub-optimal}. 
Those methods either approximate motion model or wheel odometer measurements at the cost of accuracy reduction~\cite{zhang2019large}, or rely on other \rev{complementary} sensors, e.g., an IMU~\cite{yi2007imu}. 
The latter type of methods typically requires downgrading (by sub-sampling or partially removing) the usage of wheel odometer measurements, which will also lead to information loss and eventual accuracy loss. To the best of the authors' knowledge, the problem of fully probabilistically using wheel odometer measurements for high-accuracy 6D pose estimation remains unsolved.

To this end, in this paper, 
we design pose estimation algorithmic modules dedicated to ground robots by considering both the motion and sensor characteristics. In addition, our methods are fully probabilistic and generally applicable, which can be integrated into different pose estimation frameworks. 
To achieve our goal, the key design factors and contributions are as follows.
Firstly, we propose a parametric representation method for the motion manifold on which the ground robot moves. \rev{Specifically, we choose to use second-order polynomial equations since lower-order representation is unable to represent 6D motion.}  Secondly, we design a method for performing manifold based 6D pose integration with wheel odometer measurements in closed form, without \rev{extra} approximation and information reduction. In addition, we analyze the \rev{pose estimation} errors caused by the manifold representation and present an approach for re-parameterizing manifold equations periodically \rev{for the sake of estimation accuracy}.
Moreover, we propose a complete pose estimation system inside on a manifold-assisted sliding-window estimator, which is tailored for ground robots by fully exploiting the manifold constraints and fusing measurements from both wheel odometer and a monocular camera. Measurements from an IMU can also be optionally integrated into the proposed algorithm to further improve the accuracy.

\rev{To demonstrate the effectiveness of our method,
we conducted extensive simulated and real-world experiments on multiple platforms equipped with the wheel odometer and a monocular camera}\footnote{\rev{Other exteroceptive sensors, e.g., 3D laser range-finders, can also be used in combination with the wheel odometer via the proposed method. However, providing detailed estimator formulation and experimental results using alternative exteroceptive sensors is beyond the scope of this work.}}. Our results show that the proposed method outperforms other state-of-the-art algorithms \rev{in pose estimation accuracy}, specifically~\cite{li2013high,qin2017vins,wu2017vins,zhang2019large}, by a significant margin.

\section{Related Work}
In this work, we focus on pose estimation for ground robots by probabilistically utilizing manifold constraints. 
Therefore, we group the related work into three categories: camera-based pose estimation, pose estimation for ground robots, and physical constraints assisted methods.

\subsection{Pose Estimation using Cameras}
In general, there are two families of camera-based pose estimation algorithms: the ones that rely on cameras only~\cite{DongSi2012,engel2014lsd,engel2017direct,mur2017orb} or the ones which fuse measurements from both cameras and other sensors~\cite{li2013high,forster2017manifold,schneider2018maplab,leutenegger2015keyframe}. Typically, camera-only methods require building a local map incrementally, and computing camera poses by minimizing the errors computed from projecting the local map onto new incoming images. The errors used for optimization can be either under geometrical form~\cite{DongSi2012,mur2017orb} or photometric form~\cite{engel2014lsd,engel2017direct}. On the other hand, cameras can also be used in combination with other types of sensors for pose estimation, and the common choices include IMU~\cite{li2013high,forster2017manifold,eckenhoff2019closed}, GPS~\cite{chen2018integration}, and laser range finders~\cite{zhang2015visual}. Once other sensors are used for aiding camera-based pose estimation, the step of building a local map becomes not necessary since pose to pose prior estimates can be computed with other sensors~\cite{li2013high,forster2017manifold,zhang2015visual}. Based on that, computationally light-weighted estimators can be formulated by partially or completely marginalizing all visual features to generate stochastic constraints~\cite{li2013high,li2014online}.

In terms of estimator design for camera-based pose estimation, there are three popular types: filter-based methods~\cite{li2012vision,li2013high,li2014online}, iterative optimization-based
methods~\cite{leutenegger2015keyframe,engel2017direct,schneider2017visual}, and finally learning-based methods~\cite{clark2017vidloc,bloesch2018codeslam}. The filter-based methods are typically used in computationally-constrained platforms for achieving real-time pose estimation~\cite{li2012vision}. To further improve the pose estimation accuracy, recent work~\cite{engel2017direct,leutenegger2015keyframe} introduced iterative optimization-based methods to re-linearize states used for computing both residual vectors and Jacobian matrices. By doing this, linearization errors can be reduced, and the final estimation accuracy can be improved. Inspired by recent success in designing deep neural networks for image classification~\cite{russakovsky2015imagenet}, learning-based methods for pose estimation are also under active research and development~\cite{clark2017vidloc,bloesch2018codeslam}, which in general seek to learn scene geometry representation instead of \rev{relying on} an explicit parametric formulation. 

\subsection{Pose Estimation for Ground Robots}
Pose estimation for ground robots has been under active research for the past decades~\cite{nister2006visual,Zheng2018constraint,Chenyang2018monocular,zhang2018laser,hess2016real, Zuo2019ISRR}. 
\rev{The work of} \cite{nister2006visual} is one of the well-known methods, which uses stereo camera for ground robot \rev{pose estimation}. 
For ground robot pose estimation, 
laser range finders (LRF) are also 
widely used. 
A number of algorithms were proposed by using different types of LRFs~\cite{levinson2011towards,zhang2018laser,hess2016real,Zuo2019IROS}, e.g., a
3D LRF~\cite{levinson2011towards,Zuo2019IROS}, a 2D LRF~\cite{hess2016real}, or a self-rotating 2D LRF~\cite{zhang2018laser}.
Additionally, radar sensors are of small size and low cost, and thus also suitable for a wide range of autonomous navigation tasks~\cite{checchin2010radar, adams2012robotic, ward2016vehicle}.

Similar to other tasks, learning-based methods were also proposed for ground robots. 
Chen et al.~\cite{changhao2018ionet} proposed a bi-directional LSTM deep neural network
for IMU integration by assuming zero change along the z-axis.
This method is able to \rev{estimate the pose of} ground robot in indoor environments with higher accuracy compared to traditional IMU integration. 
In~\cite{brossard2019rins}, an inertial aided pose estimation method for wheeled robots is introduced, which relies on detecting situations of interests (e.g., zero velocity event) and utilizing an invariant extended Kalman filter~\cite{barrau2016invariant} for state estimation.
\rev{Lu et al.}~\cite{Chenyang2018monocular} proposed an approach to compute a 2-dimensional semantic occupancy grid map directly from front-view RGB images, via a variational encoder-decoder neural network. Compared to the traditional approach, this method does not need the use of an IMU to compute roll and pitch and project front-view segmentation results onto a plane. Roll and pitch information is calculated implicitly from the network.

In recent years, there are a couple of low-cost pose estimation methods designed for ground robots by incorporating the usage of wheel odometer measurements~\cite{censi2013simultaneous,wu2017vins,quan2018tightly}. Specifically, 
Wu et al.~\cite{wu2017vins} proposed to introduce planar-motion constraints to \rev{visual-inertial} \rev{pose estimation} system, and also add wheel odometer measurements for stochastic optimization. The proposed method is shown to improve overall performance in {\em indoor} environments. Similarly, 
Quan et al.~\cite{quan2018tightly} designed a complete framework for visual-odometer SLAM, in which IMU measurements and wheel odometer measurements were used together for pose integration.
Additionally, to better utilize wheel odometer measurements, the intrinsic parameters of wheel odometer can also be calibrated online for performance enhancement
~\cite{censi2013simultaneous}.
However,~\cite{censi2013simultaneous,wu2017vins,quan2018tightly} only focus on robotic navigation on {\em a single piece of planar surface}. While this is typically true for most indoor environments, applying those algorithms in \rev{complex} outdoor 3D environments (e.g., urban streets) is highly risky. 

To enable 6D pose estimation for ground robots, 
\rev{Lee et al.}~\cite{lee2015online} proposed to estimate robot poses as well as ground plane parameters jointly. The ground plane is modelled by polynomial parameters, and the estimation is performed by classifying the ground region from images and using sparse geometric points within those regions. However, this algorithm is {\em not} probabilistically formulated, \rev{which might cause accuracy reduction}. Our previous work~\cite{zhang2019large} went to a similar direction by modeling the manifold parameters, while with an approximate maximum-a-posteriori estimator. Specifically, we modelled the manifold parameters as part of the state vector and used an iterative optimization-based sliding-window estimator for \rev{pose estimation}~\cite{zhang2019large}. However, to perform 6D pose integration, the work of~\cite{zhang2019large} requires using an IMU and approximating wheel odometer integration equations, which will inevitably result in information loss and accuracy reduction. In this paper, we significantly extend the work of~\cite{zhang2019large}, by introducing manifold based probabilistic pose integration in closed form, removing the mandatory needs of using an IMU, and formulating manifold re-parameterization equations. We show that the proposed algorithm achieves significantly better performance compared to our previous work and other competing state-of-the-art vision-based pose estimation methods \rev{for ground robots}. 

\subsection{Physical Constraints Assisted Methods}
In addition, since robots are deployed and tested in real environments and the corresponding physical quantities vary along with applications and scenarios, additional constraints can also be used for enhancing estimation performance. Representative \rev{constraints can be derived from} water pressure~\cite{corke2007experiments,shkurti2011state}, air pressure~\cite{engel2014scale,chen2020obtaining}, contact force~\cite{hartley2020contact}, propulsion force~\cite{nisar2019vimo} and so on.

To utilize the physical constraints, there are two main types of algorithms: by using sensors to directly measure the corresponding \rev{physical quantities}~\cite{corke2007experiments,shkurti2011state,engel2014scale,chen2020obtaining,hartley2020contact} and by indirectly formulating cost functions without sensing capabilities~\cite{nisar2019vimo}. On one hand, measurements from sensors (e.g., pressure sensors) can be used to formulate probabilistic equations by relating system state and measured values, which can be integrated into a sequential Bayesian estimation framework as extra terms~\cite{shkurti2011state,engel2014scale,hartley2020contact}. On the other hand, if a physical model is given while sensing capability is not complete, one can also explicitly consider the corresponding uncertainties to allow accurate estimation~\cite{nisar2019vimo}. \rev{The manifold constraint used in this paper belongs to the second category since the motion manifold is not directly measurable by sensors.}

\section{Notations and Sensor Models}

\begin{figure}[t]
	\centering
	\includegraphics[width=1.0\columnwidth]{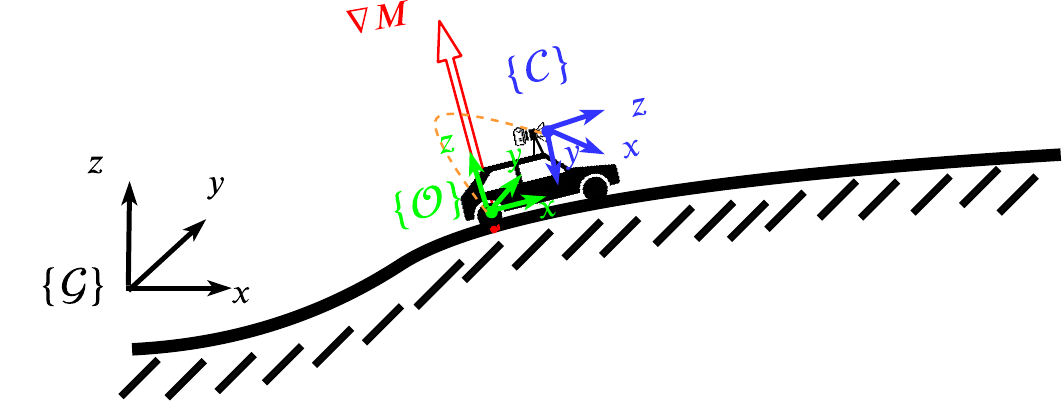} 
	
	\caption{
		Conceptual representation of a ground robot moving on a manifold. The global reference frame $\{\mathcal{G}\}$, wheel odometer frame $\{\mathcal{O}\}$, and the camera frame $\{\mathcal{C}\}$ are in black, green and blue colors, respectively. The manifold gradient vector $\nabla \mathbf{M}$ is also shown in red.
	}
	\label{fig:robot}
\end{figure}

\subsection{Notations}
In this work, we assume a ground robot navigating with respect to a global reference frame, $\{ \mathcal G \}$, whose wheels are always in contact with the road surface. We use $\{ \mathcal O \}$ to denote the wheel odometer reference frames. \rev{The reference frame of the exteroceptive sensor, i.e. a monocular camera, is denoted by $\{ \mathcal C \}$.} The center of frame $\{ \mathcal O \}$ locates
at the center of the robot wheels, with its x-axis pointing forward and z-axis pointing up (see Fig.~\ref{fig:robot} for details).
Additionally, we use $^{\mathbf A} \mathbf p _{\mathbf B}$ and $^{\mathbf A} _{\mathbf B} \bar{\mathbf q} $ to represent the position and unit quaternion orientation of frame $\mathcal B$ with respect to the frame $\mathcal A$.
$^{\mathbf A} _{\mathbf B} \bR$ is the rotation matrix corresponding to $^{\mathbf A} _{\mathbf B} \bar{\mathbf q}$. We use $\hat{\mathbf a}$,  $\tilde{\mathbf a}$, $\mathbf a^T$, $\dot{\mathbf a}$, and $||\mathbf a||$ to represent the estimate, error, transpose, time derivative, and Euclidean norm of the variable $\mathbf a$. Finally, $\mathbf e_i$ is a $3\times1$ vector, with the $i$th element to be $1$ and other elements to be $0$, \rev{and $\mathbf e_{ij} = [	\mathbf e_i, \mathbf e_j] \in \mathbb{R}^{3\times 2}$}.

\subsection{Wheel Odometer Measurement Model}
Similar to~\cite{yap2011particle,censi2013simultaneous,wu2017vins}, at time $t$, the measurements for an intrinsically calibrated wheel odometer system are given by:
\begin{align}
\label{eq:odom meas}
\mathbf u_o (t)
= 
\begin{bmatrix}
v_o (t) \\ \omega_o (t)
\end{bmatrix}
=
\begin{bmatrix}
{\mathbf e_1^T } \cdot {}^{\mathbf O(t)} \mathbf v +  n_{vo}\\ 
{\mathbf e_3^T } \cdot {}^{\mathbf O(t)} \boldsymbol \omega + n_{\omega o}
\end{bmatrix}
\end{align}
where ${}^{\mathbf O(t)} \mathbf v$ and ${}^{\mathbf O(t)} \boldsymbol \omega$ are the linear velocity and 
rotational velocity of the center of frame $\mathcal O$ expressed in the frame $\mathcal O$ at time $t$, 
and $n_{vo}$ and $n_{\omega o}$ are \rev{the white noises in measurements},  whose vector forms are 
$\mathbf n_{v} = \begin{bmatrix}
n_{vo} & 0& 0 \end{bmatrix}^\top$ and $\mathbf n_{\omega} =  \begin{bmatrix}
0& 0& n_{\omega o} \end{bmatrix}^\top$. Eq.~\eqref{eq:odom meas} clearly demonstrates that wheel odometer measurements only provide 2D motion information, i.e., forward linear velocity and rotational velocity about yaw. Therefore, by using measurements only from Eq.~\eqref{eq:odom meas}, it is theoretically possible to conduct planar-surface based pose integration, while infeasible to perform 6D pose integration.

\section{Methodology with Motion Manifold}
\label{sec: 6d mani inte}
\subsection{Manifold Representation and Induced 6D Integration}
\label{sec:mani raw}
\subsubsection{Mathematical Representation of Motion Manifold}
In order to utilize wheel odometer measurements for 6D pose integration, 
\rev{the motion manifolds where ground robots navigate} need to be mathematically modeled and integrated into the propagation process.
To this end, we model the motion manifold by parametric equations. Specifically,  
we choose to approximate the motion manifold around any 3D location $\mathbf p$ by a quadratic polynomial:
\begin{align}
	\label{eq:manifold}
	\mathcal{M}(\mathbf p) = z + 
	c + \mathbf B^T
	\begin{bmatrix}
		x \\ y
	\end{bmatrix}
	+ 
	\frac{1}{2}
	\begin{bmatrix}
		x \\ y
	\end{bmatrix}^T
	\mathbf A
	\begin{bmatrix}
		x \\ y
	\end{bmatrix} = 0,
	\mathbf p = 
	\begin{bmatrix}
		x \\ y \\ z
	\end{bmatrix}
\end{align}
with
\begin{align}
	\label{eq:man para0}
	\mathbf B = \begin{bmatrix}
		b_1 \\ b_2
	\end{bmatrix},\,
	\mathbf A = \begin{bmatrix}
		a_1 & a_2 \\
		a_2 & a_3
	\end{bmatrix}.
\end{align}

The manifold parameters are:
\begin{align}
	\label{eq:man para}
	\mathbf{m} = 
	\begin{bmatrix}
		c& b_1 & b_2 & a_1 & a_2 & a_3 
	\end{bmatrix}^\top.
\end{align}

We note that traditional methods~\cite{yap2011particle,censi2013simultaneous,wu2017vins} that assume planar-surface environments are
mathematically equivalent to model the manifold by parameter $\mathbf{m} = [c,0,0,0,0,0]^T$ or $\mathbf{m} = [c,b_1,b_2,0,0,0]^T$. Their design choices fail to represent the general condition of the outdoor road surface, and thus not suitable for high-accuracy estimation in \rev{complex} large-scale environments. It is important to point out that compared to the zeroth or first order representation, the second order one we use in Eq.~\eqref{eq:manifold} is not to simply add parameters. In fact, only second or higher order representation is able to allow 6D motion. If zeroth or first order representation is used, motion along the normal direction of the surface can never be properly characterized, leading to reduced precision. To show this statistically, 
different choices of polynomial representation are also compared in the experimental section (see Sec.~\ref{sec:simu exp}).

\subsubsection{Analysis On 6D Integration}
To date, existing methods on probabilistically consuming wheel odometer measurements can only perform 3D pose estimation~\cite{censi2013simultaneous,wu2017vins}, and it is difficult to extend the capability into 6D space. This is primarily due to the fact that odometer only provides 2D measurement (see Eq.~\eqref{eq:odom meas}).  
However, with the manifold representation defined, i.e., Eq.~\eqref{eq:man para}, 6D pose integration becomes feasible. 
To show this, we first analyze the conditions required for 6D pose integration, and subsequently discuss the intuitive motivation of our method. In the next sub-section, detailed derivations of our method is provided.

To perform 6D pose integration in an estimator, the corresponding derivative terms, i.e., rotational and linear velocities, need to be represented by functions of estimator states and sensor measurements. For example, in IMU based integration, rotational velocity is provided by gyroscope measurements, and linear velocity can be propagated by estimator state and accelerometer measurements~\cite{li2013optimization}. Therefore, to allow 6D integration, it is required to formulate 6 independent constraints from both estimator state and sensor measurements, for rotational and linear velocities respectively.

As we have mentioned, the odometer measurements, i.e. Eq.~\eqref{eq:odom meas}, explicitly provide 2 independent constraints for rotational and linear velocities respectively. To seek for the other 4 additional equations, we first utilize the non-holonomic constraint\footnote{Robots equipped with Mecanum wheels are not under this constraint, and the process must be re-designed. However, this type of wheels is not commonly used in low-cost commercial outdoor robots and vehicles.}, which \rev{eliminates} two additional degrees of freedom (DoF) on linear velocity term. Under non-holonomic constraint, local velocity of a robot is only non-zero in its x-axis, i.e. $\mathbf e^T_2 \cdot {}^{\mathbf O}\mathbf v=0$ and $\mathbf e^T_3 \cdot {}^{\mathbf O}\mathbf v=0$. The remaining required 2 DoFs are those for rotational velocity, which we find are implicitly represented by the motion manifold. Specifically, as shown by Fig.~\ref{fig:robot}, the motion manifold of each location determines roll and pitch angles of the ground robot (by the manifold gradient), and manifold across space defines how those angles evolve. Therefore, by combining odometer measurements, non-holonomic model, and the manifold representation, 6D pose integration can be performed.

\subsubsection{Manifold-Aided Integration}
In this section, we present the detailed equations for pose integration. To begin with, we first note that the kinematic equation for orientation term can be expressed by:
\begin{align}
	\label{eq:rot dir}
	{^{\bG}_{\bO(t)} \dot{\bR}} &= 
	{^{\bG}_{\bO(t)} {\bR}}  \cdot
	\lfloor {^{\bO(t)} \bomega}  \rfloor
\end{align}
where
\begin{align}
	\lfloor\boldsymbol{\omega} \rfloor=\left[\begin{array}{ccc}{0} & {-\omega_{z}} & {\omega_{y}} \\ {\omega_{z}} & {0} & {-\omega_{x}} \\ {-\omega_{y}} & {\omega_{x}} & {0}\end{array}\right],\,\,\,
	\boldsymbol{\omega} = \begin{bmatrix}
		{\omega_{x}} \\{\omega_{y}}\\{\omega_{z}}
	\end{bmatrix}
\end{align}
To perform orientation integration, ${^{\bO(t)} \bomega}$ needs to be known. However, 
it is clear that the odometer measurement only provides the third element of ${^{\bO(t)} \bomega}$, as described in Eq.~\eqref{eq:odom meas}. To obtain the first two elements of ${^{\bO(t)} \bomega}$, 
information about the manifold representation needs to be used. To derive the equations,
we first write:
\begin{align}
	\label{eq:int1}
	\big{\lfloor}
	\big{(} {^{\mathbf G}_{\mathbf O(t)} \mathbf R} \cdot {\mathbf e_3}
	\big{)} 
	\big{\rfloor}_{12} \cdot 
	\nabla \mathcal{M}
	\left( {^\bG \mathbf p _{\bO(t)}} \right) 
	= \mathbf 0
\end{align}
where $\lfloor {}\mathbf a \rfloor_{12}$ are the first two rows of the cross-product matrix $\lfloor {}\mathbf a \rfloor$ 
The above equation reveals the fact that, the motion manifold $\mathcal M$ has explicitly defined roll and pitch of a ground robot, which should be consistent with the rotation matrix
$^{\mathbf G}_{\mathbf{O(t)}} \mathbf R$. In other words, the manifold gradient vector should be \rev{parallel to} $^{\mathbf G} \mathbf z_{\mathbf{O}}$, 
	$^{\mathbf G} \mathbf z_{\mathbf{O}} = {^{\mathbf G}_{\mathbf O(t)} \mathbf R} \cdot {\mathbf e_3}$ (see Fig.~\ref{fig:robot}).
Taking time derivative of Eq.~\eqref{eq:int1} leads to:
\begin{align}
	\label{eq:int2}
	\big{\lfloor} {^{\mathbf G}_{\mathbf O(t)} \dot{\bR}}  {\mathbf e_3}
	\big{\rfloor}  
	\nabla \mathcal{M}
	\left( {^\bG \mathbf p _{\bO(t)}} \right) \! +\!
	\big{\lfloor}
	{^{\mathbf G}_{\mathbf O(t)} 
		{\bR}}  {\mathbf e_3}
	\big{\rfloor} 
	\dot{\nabla \mathcal{M}}
	\left( {^\bG \mathbf p _{\bO(t)}} \right)
	= \mathbf 0    
\end{align}
By using $\mathcal{M}(t) = \mathcal{M}
\left( {^\bG \mathbf p _{\bO(t)}} \right) $ for abbreviation, and substituting Eq.~\eqref{eq:int1} into Eq.~\eqref{eq:int2}, we obtain:
\begin{align}
	\label{eq:int3}
	&\big{\lfloor}
	{^{\mathbf G}_{\mathbf O(t)} \bR 
		\lfloor {^{\bO(t)}\bomega }  \rfloor} {\mathbf e_3} 
	\big{\rfloor}  
	\nabla \mathcal{M} (t) +
	\big{\lfloor}
	{^{\mathbf G}_{\mathbf O(t)} 
		{\bR}} \cdot {\mathbf e_3}
	\big{\rfloor} \textsf{}
	\dot{\nabla \mathcal{M}} (t)
	= \mathbf 0 \\
	\Rightarrow & 
	\big{\lfloor}
	{^{\mathbf G}_{\mathbf O(t)} \bR^T \nabla \mathcal{M} (t) } 
	\big{\rfloor} 
	\lfloor {^{\bO(t)}\bomega }  \rfloor
	{\mathbf e_3}
	+
	\big{\lfloor}
	{^{\mathbf G}_{\mathbf O(t)} 
		{\bR}^T} \dot{\nabla \mathcal{M}} (t)
	\big{\rfloor} {\mathbf e_3}
	= \mathbf 0 \\
	\Rightarrow &
	\label{eq:int4}
	\big{\lfloor}
	{^{\mathbf G}_{\mathbf O(t)} \bR^T \nabla \mathcal{M} (t) }  
	\big{\rfloor} 
	\lfloor {\mathbf e_3} \rfloor
	{^{\bO(t)}\bomega } 
	=
	\big{\lfloor}
	{^{\mathbf G}_{\mathbf O(t)} 
		{\bR}^T} \dot{\nabla \mathcal{M}} (t)
	\big{\rfloor} {\mathbf e_3}
\end{align}
Eq.~\eqref{eq:int4} contains 3 linear equations for ${^{\bO(t)}\bomega }$, in which however the number of linearly independent equations is only 2. Specifically, the un-identified variable corresponds to the third element in ${^{\bO(t)}\bomega }$, 
since ${^{\bO(t)}\bomega }$ is left multiplied by $\lfloor {\mathbf e_3} \rfloor$:
\begin{align}
		\lfloor {\mathbf e_3} \rfloor = 
		\begin{bmatrix}
			0 & -1 & 0 \\
			1 & 0 & 0 \\
			0 & 0 & 0
		\end{bmatrix}, 
\end{align}On the other hand, 
the third element in ${^{\bO(t)}\bomega }$ can be directly available from the wheel odometer measurement (see Eq.~\eqref{eq:odom meas}). Those observations verify our design motivation of combining wheel odometer measurement and manifold representation together, to rely on their complementary properties for 
performing 6D pose integration. 

To complete our derivation, we seek to obtain the first and second elements of ${^{\bO(t)}\bomega }$ from Eq.~\eqref{eq:int4}.
To do this, we first write the following equation based on Eq.~\eqref{eq:int1}:
\begin{align}
	\nabla \mathcal{M}(t) = 
	\left\| \nabla \mathcal{M}(t) \right\| \cdot 
	^{\bG}_{\bO (t)} \bR \cdot \mathbf e_3
\end{align}
As a result, Eq.~\eqref{eq:int4} becomes:
\begin{align}
	\label{eq:int5}
	& \left\| \nabla \mathcal{M}(t) \right\|
	\lfloor {\mathbf e_3} \rfloor
	\lfloor {\mathbf e_3} \rfloor
	{^{\bO(t)}\bomega } 
	=
	\big{\lfloor}
	\big{(} {^{\mathbf G}_{\mathbf O(t)} 
		{\bR}^T} \cdot \dot{\nabla \mathcal{M}} (t)
	\big{)} 
	\big{\rfloor} {\mathbf e_3} \\
	\label{eq:int6}
	\Rightarrow &
	\left\| \nabla \mathcal{M}(t) \right\|
	{^{\bO(t)}\bomega _{12}} 
	=
	\mathbf e^T_{12}
	\big{\lfloor}
	{\mathbf e_3}
	\big{\rfloor}
	\big{(} {^{\mathbf G}_{\mathbf O(t)} 
		{\bR}^T} \cdot \dot{\nabla \mathcal{M}} (t)
	\big{)} \\
	\label{eq:int7}
	\Rightarrow &
	{^{\bO(t)}\bomega _{12}} 
	=
	\frac{1}{\left\| \nabla \mathcal{M}(t) \right\|}
	\mathbf e^T_{12}
	\big{\lfloor}
	{\mathbf e_3}
	\big{\rfloor}
	\big{(} {^{\mathbf G}_{\mathbf O(t)} 
		{\bR}^T} \cdot \dot{\nabla \mathcal{M}} (t)
	\big{)}
\end{align}
where 
${^{\bO(t)}\bomega _{12}} = \mathbf e^T_{12} {^{\bO(t)}\bomega}$.
We note that, in Eq.~\eqref{eq:int5}, we have used the equality:
\begin{align}
		\big{\lfloor}
		{\mathbf e_3}
		\big{\rfloor}
		\big{\lfloor}
		{\mathbf e_3}
		\big{\rfloor} = 
		-\begin{bmatrix}
			1 & 0 &  0\\
			0 & 1 & 0 \\ 
			0 & 0 & 0
		\end{bmatrix}
\end{align}By considering odometer measurements, we have:
\begin{align}
	\label{eq: final omega}
	^{\bO(t)}\bomega = 
	\begin{bmatrix}
		\frac{1}{\left\| \nabla \mathcal{M}(t) \right\|}
		\mathbf e^T_{12}
		\big{\lfloor}
		{\mathbf e_3}
		\big{\rfloor}
		\big{(} {^{\mathbf G}_{\mathbf O(t)} 
			{\bR}^T} \cdot \dot{\nabla \mathcal{M}} (t)
		\big{)} \\
		\omega_o (t) - 
		n_{\omega o}
	\end{bmatrix}
\end{align}
By integrating Eq.~\eqref{eq: final omega}, 3D orientation estimates can be computed.
Once orientation is computed, we compute position by integrating:
\begin{align}
		\label{eq: final v}
		^\bG \dot{\bp} _{\bO(t)} &= 
		{} ^\bG {\bv} _{\bO(t)} = 
		{^{\mathbf G}_{\mathbf O(t)} }
		{\bR} \cdot
		{} ^{\bO(t)} {\bv},\,
		{} ^{\bO(t)} {\bv} = 
		\begin{bmatrix}
			v_o - n_{v o} \\   0 \\ 0
		\end{bmatrix}
\end{align}
It is important to point out that the above equation is formulated by combining the information provided from both the wheel odometer measurement (first row of $^{\bO(t)} {\bv}$) and non-holonomic constraint (second and third row of $^{\bO(t)} {\bv} $).

We also note that our manifold representation, i.e.,~Eq.~\eqref{eq:manifold}, implicitly defines a motion model 
that the integrated position must satisfy, $\mathcal{M}(^\bG {\bp} _{\bO(t)}) = 0$. 
Actually, the representation of Eq.~\eqref{eq: final v} exactly \rev{satisfies} the manifold constraint. 
To show the details, we note that:
\begin{align}
	\mathcal{M}(^\bG {\bp} _{\bO(t)}) = 0 \Rightarrow	&
	\frac{\partial \mathcal{M}(^\bG {\bp} _{\bO(t)})}{\partial t} = 0 \notag \\
	\Rightarrow  & 
	\nabla \mathcal{M}(^\bG {\bp} _{\bO(t)})^T \cdot {^\bG \bv _{\bO(t)}} = 0 
\end{align}
Since $^{\bO(t)}_\bG \bR^T\mathbf e_3$ is collinear with $\nabla \mathcal{M}(^\bG {\bp} _{\bO(t)})$, it leads to:
\begin{align}
	\label{eq:v verify}
	\mathbf e_3^T 
	\cdot {}
	^{\bO(t)}_\bG \bR 
	\cdot 
	{^\bG \bv _{\bO(t)}} = 0
\end{align}
\rev{The expression of ${^\bG \bv _{\bO(t)}}$ in Eq.~\eqref{eq: final v} clearly satisfies Eq.~\eqref{eq:v verify} deduced from the manifold constraint}.

With Eq.~\eqref{eq: final omega} and~\eqref{eq: final v} being defined, we are able to perform odometer based 6D pose integration on a manifold. The entire process does not include an IMU. Instead, by combining the information provided by the odometer measurement, non-holonomic model, and manifold equations, the integration requirement is fulfilled. We emphasize that our method allows the use of odometer measurements for probabilistically 6D integration, which is {\em not} feasible in previous literature.

\subsection{State and Error-State Prediction}
\label{sec: manifold formal int}
In this section, we describe the details of using the proposed manifold representation for performing state and error-state propagation, which is a necessary prerequisite for formulating probabilistic estimators~\cite{trawny2005indirect}.
Since the integration process requires explicit representations of 
both poses (position and orientation vectors) and manifold parameters, we define the state vector as follows:
\begin{align}
	\label{eq:small x}
	\bst = \begin{bmatrix}
		{}^{\bG}\bp_{\bO}^\top & {}^{\bG}_{\bO}\bq^\top & \bm^\top
	\end{bmatrix}^\top
\end{align}
The time evolution of the state vector can be described by:
\begin{align}
	\label{eq:continous evolution0}
	{}^{\bG}\dot{\bp}_{\bO(t)} &= {}^{\bG}_{\bO(t)}\bR {}^{\bO(t)}\bv\\
	{}^{\bG}_{\bO(t)}\dot{\bq} &= \frac{1}{2} {}^{\bG}_{\bO(t)}{\bq} \cdot \boldsymbol{\Omega}(\boldsymbol{{}^{\bO(t)}\omega}) \\
	\dot{\bm(t)} &=  \mathbf 0
	\label{eq:continous evolution}
\end{align}
where 
\begin{align}
	\boldsymbol{\Omega}(\boldsymbol{\omega})=\left[\begin{array}{cc}{-\lfloor\boldsymbol{\omega} \rfloor} & {\boldsymbol{\omega}} \\ {-\boldsymbol{\omega}^{T}} & {0}\end{array}\right]
\end{align}
We point out that, during the prediction stage, our previous work~\cite{zhang2019large} utilizes $\dot{\bm(t)} =  \mathbf n_{wm}$ instead of Eq.~\eqref{eq:continous evolution} for describing the kinematics of the motion manifold, where $\mathbf n_{wm}$ is a zero mean Gaussian vector. 
The design of $\dot{\bm(t)} =  \mathbf n_{wm}$ is under the consideration that time prediction is typically performed for small time windows in which well-constructed road surfaces are smooth and of slow changes. Those changes can be captured via uncertainty propagation by $\mathbf n_{wm}$.  
\rev{However, the characterization of $\dot{\bm(t)} =  \mathbf n_{wm}$ becomes inaccurate sometimes, especially in complex environments.}
To ensure robust pose  estimation, we utilize Eq.~\eqref{eq:continous evolution} and additional algorithmic modules, which are illustrated in details in Sec.~\ref{sec:re-param}.
Based on Eqs.~\eqref{eq:continous evolution0}-\eqref{eq:continous evolution}, the dynamics of the estimated poses can be written as:
\begin{align}
	\label{eq:c_est0}
	{}^{\bG}\hat{\dot{\bp}}_{\bO(t)} &= {}^{\bG}_{\bO(t)}\hat{\bR} {}
	\begin{bmatrix}
		v_o(t) & 0 & 0 
	\end{bmatrix}^T \\
	{}^{\bG}_{\bO(t)}\hat{\dot{\bq}} &= \frac{1}{2} {}^{\bG}_{\bO(t)}\hat{\bq} \cdot \boldsymbol{\Omega}(\boldsymbol{{}^{\bO(t)}\hat{\omega}}) \\
	\label{eq:c_est2}
	\hat{\dot{\bm(t)}} &=  \mathbf 0
\end{align}
where
\begin{align}
	\label{eq: omega error1}
	^{\bO(t)}
	\hat{\bomega} = 
	\begin{bmatrix}
		\frac{1}{\left\| \nabla \hat{\mathcal{M}}(t) \right\|}
		\mathbf e^T_{12}
		\big{\lfloor}
		{\mathbf e_3}
		\big{\rfloor}
		\big{(} {^{\mathbf G}_{\mathbf O(t)} 
			\hat{\bR}^T} \cdot \hat{\dot{\nabla \mathcal{M}}} (t)
		\big{)} \\
		\omega_o (t)
	\end{bmatrix}
\end{align}
Eqs.~\eqref{eq:c_est0} to~\eqref{eq:c_est2} can be integrated with 
wheel odometer measurement via numerical integration methods (e.g., Runge-Kutta).

To describe the details of error-state propagation, we first 
employ first order approximation of \rev{Taylor} series expansion on Eq.~\eqref{eq: final omega}, to obtain:
\begin{align}
	\label{eq: omega error}
	^{\bO(t)}\bomega = {}
	^{\bO(t)}\hat{\bomega} + {}
	^{\bO(t)}\tilde{\bomega}
	={}
	^{\bO(t)}\hat{\bomega} +
	\mathbf J_\bst \bstt
	+
	\mathbf J_\bnn \mathbf{n}
\end{align}
where
\begin{align}
	\bn = \begin{bmatrix}
		n_{v_o} & n_{\omega_o} 
	\end{bmatrix}^\top 
\end{align}
and
\begin{align}
	\mathbf J_\bst = 
	\frac{\partial{^{\bO(t)}
			{\bomega}}} 
	{\partial{\bst}},\,\,\,\,\,
	\mathbf J_\bnn = 
	\frac{\partial{^{\bO(t)}
			{\bomega}}} 
	{\partial{\bnn}}
\end{align}
To derive orientation error-state equation, we first define orientation error vector. Specifically, we employ small error approximation to represent a local orientation error $\delta \btheta$ by~\cite{trawny2005indirect}:
\begin{align}
	\label{eq:rot error}
	^\bO _\bG \bR \simeq
	\left( {\mathbf I - \lfloor \delta \btheta
		\rfloor} \right) {}
	^\bO _\bG \hat{\bR}
\end{align}
By linearizing Eqs.~\eqref{eq:c_est0} to~\eqref{eq:c_est2}, we are able to write:
\begin{align}
	\label{eq:derivative theta}
	\dot{\delta \btheta} = 
	-\lfloor {^{\bO(t)}
		\hat{\bomega}} \rfloor
	\delta \btheta  +
	\mathbf J_\bst \bstt
	+
	\mathbf J_\bnn \mathbf{n}
\end{align}
The detailed derivation of Eq.~\eqref{eq:derivative theta} is shown in Appendix~\ref{sec:orentation error}-\ref{sec:jacobian omega}.
Since $\delta \btheta$ is part of 
$\bstt$, the first two terms in Eq.~\eqref{eq:derivative theta} can be combined for formulating:
\begin{align}
	\label{eq:derivative theta1}
	\dot{\delta \btheta} = 
	\bar{\mathbf J}_\bst \bstt
	+ 
	\mathbf J_n \mathbf{n}
\end{align}
In terms of position, based on Eq.~\eqref{eq: final v}, it is straightforward to obtain:
\begin{align}
	\label{eq:derivative v}
	^\bG \dot{\tilde{\bp}} _{\bO(t)} &= 
	{^{\mathbf G}_{\mathbf O(t)} }
	\hat{\bR} 
	\left(
	\mathbf I +
	\lfloor {\delta \btheta} \rfloor
	\right)
	\begin{bmatrix}
		v_o -  n_{vo} \\    0 \\ 0 
	\end{bmatrix} -
	{^{\mathbf G}_{\mathbf O(t)} }
	\hat{\bR} 
	\begin{bmatrix}
		v_o\\    0 \\ 0 
	\end{bmatrix} \\
	&=
	\label{eq:derivative v1}
	- {^{\mathbf G}_{\mathbf O(t)} }
	\hat{\bR} 
	\left\lfloor 
	\begin{bmatrix}
		v_o \\ 0 \\ 0
	\end{bmatrix}
	\right\rfloor \delta \btheta - 
	{^{\mathbf G}_{\mathbf O(t)} }
	\hat{\bR} 
	\begin{bmatrix}
		n_{vo}\\    0 \\ 0 
	\end{bmatrix}
\end{align}
For the manifold parameters we have:
\begin{align}
	\dot{\tilde{\bm}} = \mathbf 0
	\label{eq:manifold error state}
\end{align}
With Eqs.~\eqref{eq:derivative theta}, \eqref{eq:derivative v1}, \eqref{eq:manifold error state} being defined, we are able to perform error-state integration in 6D. To put all equations in the vector form, we have the error state equation as:
\begin{align}
	\bstdt = \mathbf{F}_c \cdot \bstt + \bG_c \cdot \bn, \,\,\,
	\bstt = 
	\begin{bmatrix}
		\delta \btheta^T &^\bG {\tilde{\bp}} _{\bO}^T & \tilde{\bm}^T
	\end{bmatrix}^T
\end{align}
where
\begin{align}
	\mathbf{F}_c = \begin{bmatrix}
		\mathbf{0}_{3 \times3} &  -{^{\mathbf G}_{\mathbf O} } \hat{\bR} \lfloor {}^{\bO}\bv \rfloor & \mathbf{0}_{3 \times 6} \\
		\mathbf{0}_{3 \times3} & - \lfloor {}^{\bO} \hat{\bomega} \rfloor &\mathbf{0}_{3 \times6} \\
		\mathbf{0}_{6 \times 3} & \mathbf{0}_{6 \times 3} & \mathbf{0}_{6 \times 6}
	\end{bmatrix}
	+ \begin{bmatrix}
		\mathbf{0}_{3 \times 12}\\
		\bJ_\bst \\
		\mathbf{0}_{6 \times 12}
	\end{bmatrix}
\end{align}
and
\begin{align}
	\mathbf{G}_c = \begin{bmatrix}
		- \left( {^{\mathbf G}_{\mathbf O} } \hat{\bR} \right)_{[:,1]} & \mathbf{0}_{3 \times 1}   \\
		(\bJ_n)_{[:,1]} & (\bJ_n)_{[:,2]} \\
		\mathbf{0}_{6 \times 1} & \mathbf{0}_{6 \times 1} 
	\end{bmatrix}
\end{align}
with $\left( \cdot \right)_{[:,i]}$ representing the $i$th column of a matrix. We note that $\mathbf F _c$ and $\mathbf G _c$ are continuous-time error-state transition matrix and noise Jacobian matrix. To implement a discrete-time probabilistic estimator, 
discrete-time error state transition matrix $\mathbf{\Phi}(t, \tau)$ is required. 
This can be achieved by integrating
\begin{align}
	\dot{\mathbf{\Phi}}(t, \tau) = 
	\mathbf F _c(t) \mathbf{\Phi}(t, \tau), \,\,
	\mathbf{\Phi}(\tau, \tau) = 
	\mathbf I_{12\times12}
\end{align}
using numerical integration methods. The detailed implementation of numerical integration and the remaining steps are standard ones for discrete-time estimators, which can be found at~\cite{li2013high,trawny2005indirect}.

\section{Manifold Re-Parameterization}
\label{sec:re-param}
\subsection{Challenges in Manifold Parameterization}
Eq.~\eqref{eq:manifold} defines a motion manifold in global reference frame, 
which can be probabilistically integrated for pose 6D estimation.
However, further investigation reveals that this representation itself is unable to accurately capture the kinematics of manifold parameters since real-world motion manifold changes over space.~\cite{zhang2019large} simply uses $\dot{\mathbf m} = \mathbf n_{\omega m}$, with 
$\mathbf n_{\omega m}$ being 
a zero-mean Gaussian noise vector. However, this method has limitations for large-scale deployment and is unable to approximate the dynamics of complex motion manifold parameters.
\begin{figure}[t]
	\centering
	\includegraphics[width=0.8\columnwidth]{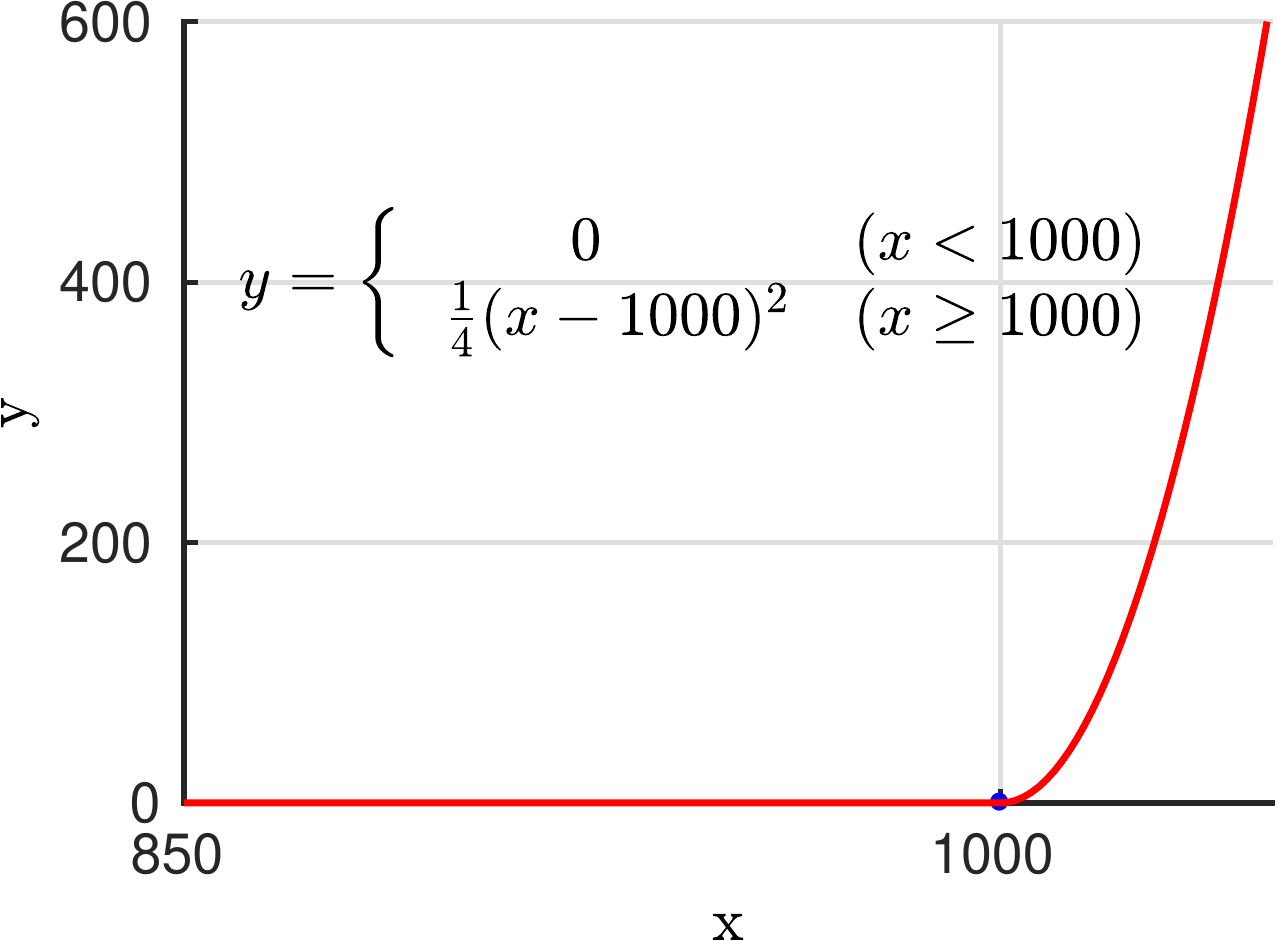} 
	
	\caption{Example of a two-dimensional manifold, concatenated by a zeroth order and a second order polynomial equation.}
	\label{fig:example mani}
\end{figure}

To identify the problem and reveal the necessity of introducing additional algorithmic design, let's look at the following equation as an example trajectory:
\begin{align}
	\label{eq:exam2d1}
	y =
	\begin{cases}
		0& {x < 1000}\\
		\frac{1}{4}(x - 1000)^2& {x \geq 1000}
	\end{cases}
\end{align}
This equation is also plotted in Fig.~\ref*{fig:example mani}. In fact, Eq.~\eqref{eq:exam2d1} can be considered as a two-dimensional manifold.
If a {\em quadratic} two-dimensional manifold is represented in general by:
\begin{align}
	\label{eq:m2}
	y = c + b x + a x^2
\end{align}
Eq.~\eqref{eq:exam2d1} is in fact 
a piecewise 
quadratic two-dimensional manifold, whose parameters in piecewise form are:
\begin{align}
	\label{eq:exam2d2}
	\begin{cases}
		c \!=\!0, b \!=\!0, a \!=\! 0& {x\! < \!1000}\\
		c \!=\! 2.5\cdot 10^5, b \!=\! -500, a \!=\! 0.25& {x \!\geq\! 1000}
	\end{cases}
\end{align}
Clearly, there will be `significant' jumps in the
manifold parameters at $x = 1000$, 
	\begin{align}
		\delta \mathbf m \!
		=\! [\delta a, \delta b, \delta c]
		\!=\! [2.5\cdot10^5,-500,0.25],\,
		|| \delta \mathbf m || \approx  2.5\cdot10^5
	\end{align}
due to our choice of quadratic representation to model a locally non-quadratic equation.
If we were able to use an infinite number of polynomials for manifold approximation, this problem should not have been caused. However, this is computationally infeasible.

In fact, to probabilistically characterize this behavior in the existing formulation, it requires the covariance of $\mathbf n_{\omega m}$, from the manifold kinematic equation $\dot{\mathbf m} = \mathbf n_{\omega m}$, to cover the significant changes of $\delta \bm$. In that case, the convergence of manifold parameter $\mathbf m$ will become problematic in stochastic estimators. If the covariance of $\mathbf n_{\omega m}$ is set to be small, it is then unable to characterize such significant changes.
%
Since manifold parameters are represented by joint probability distribution along with poses in our state vector, ill-formulated manifold parameters will inevitably lead to reduced pose estimation accuracy.

Motivated by the concept of local feature parameterization in computer vision community~\cite{civera2008inverse}, to tackle this problem, we propose to parameterize manifold equation locally.
Specifically, if Eq.~\eqref{eq:m2} is parameterized at a local point $x_o$, the corresponding equation in our example can be modified as:
\begin{align}
	\label{eq:m3}
	y = c_{new} + b_{new} ( x - x_o ) + a_{new} (x - x_o)^2
\end{align}
where $x_o$ is a fixed constant parameter. 
By applying Eq.~\eqref{eq:m3} back in the example of Fig.~\ref*{fig:example mani}, the manifold parameters in piecewise form are:
	\begin{align}
		\label{eq:exam2d22}
		\color{black}
		\begin{cases}
		c_{new} \!=\!0, \\ b_{new} \!=\!0, \\ a_{new} \!=\! 0, {x\! < \!1000}
		\end{cases}
		\begin{cases}
			c_{new} \!=\! 2.5\cdot 10^5\!+\!b_{new}x_o\!-\!a_{new}x^2_o,   \\
			b_{new} \!=\!  -500\!+\!2a_{new}x_o, \\ 
			a_{new} \!=\! 0.25, \,\,\,\,\,\,\,\,\,\,\,\,\,\,\,\,\,\,\,\,\,\,\,\,\,\,\,\,\,\,\,\,\, {x \!\geq\! 1000}
		\end{cases}
		\color{black}
	\end{align}
	which makes the `jump' in the manifold parameter estimates as functions of $x_o$:
	\begin{align}
		\delta \mathbf m
		&= [0.25(x_o-1000)^2,0.5(x_o\!-\!1000),0.25]
\end{align}
For the given example, we observe that the `jumps' in manifold parameters will be decreased if $x_o$ is close to $1000$. Assuming the manifold equation is re-parametrized at $x_o = 999.9$, we obtain the new manifold parameters
\begin{align}
	\label{eq:exam2d4}
	c_{new}  = 0.0025,\,\, b_{new} = 0.05,\,\, a_{new} =  0.25
\end{align} 
As a result, the changes of manifold parameters become much smaller and smoother.

In the given example, the manifold parameters only change at $x = 1000$, and thus parameterizing $x_o$ around $1000$ is enough.
However, when a robot traverses \rev{complex} trajectories in real scenarios, there is a great probability that the corresponding manifold parameters vary at every possible location. Therefore, it is preferable to always re-parameterize the manifold equation at the {\em latest} position. 
Additionally, since obtaining $100\%$ precise value of latest position is an infeasible task for stochastic estimators, it is a reasonable choice to use the position estimate for $x_o$, and $999.9$ is a representative estimated value in our previous example.

\subsection{Analytical Solution}
After illustrating our motivation and high-level concept by the above example, we 
introduce our formal mathematical equation of the `local' manifold representation 
and re-parameterization.
By assuming the previous re-parameterization step is performed at time $t_k$ and the next one is triggered at $t_{k+1}$, we have
\begin{align}
\label{eq:re-param}
\mathbf m (t_{k+1}) \!=\! 
\underbrace{\begin{bmatrix}
	1 & \dbp^T & \boldsymbol{\gamma}^T \\
	\mathbf 0 & \dbR^T & \boldsymbol{\Xi} \\
	\mathbf 0 & \mathbf 0 & 
	\boldsymbol{\Psi}
	\end{bmatrix}}_{\boldsymbol{\Lambda} (t_k)}
\mathbf m (t_{k}),\,\text{with}\,
\begin{array}{ll}
\dbR = \bR_o \bR_1^{-1} \\
\dbp =
\bR_o (\bp _1 \!-\! \bp_o)
\end{array}
\end{align} 
In above equation, $\bR_o$ and $\bp _o$ are the {\em fixed} re-parameterization points for $\mathbf m (t_{k})$, and  $\bR_1$ and $\bp _1$ are those for $\mathbf m (t_{k+1})$ respectively. $\boldsymbol{\gamma}$, $\boldsymbol{\Xi}$, and $\boldsymbol{\Psi}$ are all functions of $\dbR$ and $\dbp$. 
As mentioned earlier, the natural way of choosing $p_o$ and $p_1$ is to let them equal to $^\bG \bp _{\bO(t_k)}$ and $^\bG \bp _{\bO(t_{k+1})}$ respectively. Since obtaining ground truth values for those two variables is infeasible, we utilize estimated states from our estimator as replacement. Additionally, since we use sliding-window iterative estimator to compute robot poses as proposed in the next section, different estimates for the same pose will be obtained at each iteration and at each overlapped sliding window. To avoid re-parameterizing manifold state multiple times for the same pose, we pick and fix $p_o$ and $p_1$ using the state estimates at the $0$th iteration during sliding-window iterative optimization once $^\bG \bp _{\bO(t_k)}$ and $^\bG \bp _{\bO(t_{k+1})}$ firstly show up in the sliding window.
The detailed derivation of Eq.~\eqref{eq:re-param} and discussion about parameter choices of $\bR_o$ and $\bR_1$ can be found in Appendix~\ref{app:reparam-app}.

Therefore,
by employing Eq.~\eqref{eq:re-param}, we are able to
re-parameterize manifold representation around any time $t$, 
for both state estimates and their corresponding uncertainties.
We note that this is similar to re-parameterize SLAM features with inverse 
depth parameterization~\cite{li2012vision,li2013optimization}. It is important to point out that, during our re-parameterization process, $\bR_o$, $\mathbf p_o$,
$\bR_1$, $\mathbf p_1$ are regarded as {\em fixed} vectors and matrices, instead of {\em random variables}. This removes the needs of computing the corresponding Jacobians with respect to them, which is also \rev{widely used} in SLAM feature re-parameterization~\cite{li2012vision,li2013optimization}.

Once re-parameterization is introduced, the manifold stored in the state vector at time $t_{k+1}$ becomes $\mathbf m(t_{k+1})$. 
It is important to note that the manifold-based pose integration equations as described in Sec.~\ref{sec: manifold formal int} does not assume the existence of re-parameterization step, and the corresponding Jacobians are all computed with respect to the one manifold representation $\mathbf m$ (which can be termed the `global' or `initial' representation), e.g., $\frac{\partial {^{\bO}\bomega}}{\partial  \bm}$.
To allow correct prediction process, we should either derive the corresponding equations using re-parameterized manifold parameters at each instance (e.g., directly computing $\frac{\partial {^{\bO}\bomega}}{\partial  \bm(t_{k+1})}$), or utilize chain rule for calculation (e.g., computing $\frac{\partial {^{\bO}\bomega}}{\partial  \bm(t_{k+1})} = \frac{\partial {^{\bO}\bomega}}{\partial  \bm} \frac{\partial \bm}{\partial  \bm(t_{k+1})}$).
In this work, we adopt the second method, relying on relation between $\mathbf m$ and $\mathbf m(t_{k+1})$:
\begin{align}
	\mathbf m(t_{k+1}) &\!=\! \boldsymbol{\Lambda}(t_{k}) \mathbf m(t_{k}) \\ 
	&\!=\! \boldsymbol{\Lambda}(t_{k}) 
	\boldsymbol{\Lambda}(t_{k-1})\mathbf m(t_{k-1}) \!=\! \cdots \!=\!
	\prod_{i = 0}^{k} \boldsymbol{\Lambda}(t_i) \mathbf m
\end{align}
To this end, the chain rule calculation can be performed via constant matrix $\bar{\boldsymbol{\Lambda}}^{-1}_k$:
\begin{align}
		\label{eq:recover m}
		\mathbf m ={\bar{\boldsymbol{\Lambda}}^{-1}_k} \cdot \mathbf m(t_k),\,\bar{\boldsymbol{\Lambda}}_k =  \prod_{i = 0}^{k} \boldsymbol{\Lambda}(t_i)
\end{align}

With the manifold re-parameterization process defined, we revisit the 
the problem of characterizing the dynamics of the motion manifold.
Since the actual manifold on which a ground robot navigates changes over time, 
this fact must be {\em explicitly} modeled for \rev{accurate pose estimation}. Specifically, 
we take this into account during the re-parameterization process as:
\begin{align}
	\label{eq:per axis m}
	&\mathbf m (t_{k+1})  =  {\boldsymbol{\Lambda}(t_{k})} \mathbf m (t_{k}) + \bn_{wm} \\
	&\qquad \qquad \mathbf e_i^T \bn_{wm} \sim \mathcal{N}( 0, \sigma^2_{i,wm})\, \text{with}\, i = 1,\,\cdots,\,6 \notag
\end{align}
where $\mathcal{N}(0, \sigma^2_{i,wm})$ represents
Gaussian distribution with mean $0$ and variance $\sigma^2_{i,wm}$. Specifically, $\sigma _{i,wm}$ is defined by:
\begin{align}
	\label{eq:reparam_sig}
	\sigma_{i,wm} = 
	\alpha_{i,p} || \dbp ||
	\!+\! \alpha_{i,q} || \dbR ||
\end{align}
\rev{where $\alpha_{i,p}$ and $\alpha_{i,q}$ are the scalars weighting the translation and orientation.}
It is important to point out in our proposed formulation, the manifold uncertainties (i.e., $\sigma_{i,wm}$) are functions of spatial displacement as Eq.~\eqref{eq:reparam_sig}. This is different from standard noise propagation equations in VIO literatures~\cite{li2013high,forster2017manifold}, in which noises are characterized by functions of time. Since the motion manifold is a `spatial' concept instead of a `temporal' concept, this design choice better fits our estimation problem \rev{in real-world applications}.
Consequently, after the uncertainty of the motion manifold is characterized and incorporated in $\mathbf m (t_{k+1})$, the corresponding uncertainty terms are automatically converted via Eq.~\eqref{eq:recover m}.

\renewcommand{\algorithmicrequire}{\textbf{Propagation:}} 
\renewcommand{\algorithmicensure}{\textbf{Update:}} 

\begin{algorithm}
	\caption{Pipeline of the Proposed Manifold Based 6D Pose Estimation System for Ground Robots.} 
	\label{alg:Framework} 
	\begin{algorithmic}[1] 
		\Require 
		Propagate state vector using
		wheel odometer readings as well as current estimated manifold parameters (Sec.~\ref{sec: manifold formal int}). 
		\Ensure 
		When a new keyframe is allocated, 
		\State 
		Image Processing: Extracting and matching features.
		\State 
		Iterative Optimization: Performing statistical optimization using cost functions of odometry propagation, camera observations, prior term, manifold constraint (Sec.~\ref{sec:loc}). IMU  integration term can also be leveraged optionally.  
		\State 
		State Marginalization: Marginalizing estimated features, oldest keyframes, and pose integration constraints.
		If an IMU is available, we also marginalize IMU velocities and biases that are not corresponding to the latest keyframe.
		\State 
		Manifold Re-Parameterization: Re-parameterizing motion manifold using the estimated poses and also performing uncertainty propagation (Sec.~\ref{sec:re-param}).
	\end{algorithmic} 
\end{algorithm}

\section{Manifold Assisted Pose Estimation}
\label{sec:system}
The core contribution of this paper is to improve the pose estimation of the ground robot by probabilistically modeling the motion manifold and utilizing wheel odometer measurements for pose estimation. After proposing the detailed algorithmic modules in the previous section, we here demonstrate how they can be seamlessly integrated into state-of-the-art pose estimators. Specifically, we 
present a sliding-window pose estimation framework tailored for ground robots by using the proposed algorithmic modules and also fusing measurements from a monocular camera, wheel odometer, and optionally an IMU. 
The proposed method builds a state vector consisting of both motion manifold parameters and other variables of interests, performs 6D integration with the assist of the manifold, and minimizes the cost functions originating from both the raw sensor measurements and manifold constraints. The minimal sensor set required by the proposed method only includes a monocular camera and the wheel odometer.  By adding an extra IMU sensor into the proposed framework, accuracy can be further improved.

In this paper, we assume that the wheel odometer and camera share the same time clock, which is ensured by the hardware design, and all the sensors are rigidly connected to the robot. Since measurements received from different sensors are not at identical timestamps, we linearly interpolate odometer measurements (also for IMU measurements if they are available) at image timestamps.

\subsection{State Vector and Cost Functions}
\label{sec:loc}
To illustrate the 6DoF pose estimation algorithm, we first introduce the state vector.
At timestamp $t_k$, the state vector is~\footnote{For simpler representation, we ignore sensor extrinsic parameters in our presentation in this section. However, in some of our real world experiments when offline sensor extrinsic calibration is not of high accuracy, those parameters are explicitly modeled in our formulation and used in optimization.}:
\begin{align}
	\label{eq:state}
	\mathbf y_k 
	= 
	\begin{bmatrix}
		{\mathbf O_k}^T &
		\bst_k^T 
	\end{bmatrix}^T\!,
\end{align}
where $\bst$ is defined in Eq.~\eqref{eq:small x} and
${\mathbf O_k}$ \rev{denotes the poses in the sliding-window at timestamp k}:
\begin{align}
	{\mathbf O_k} \!\!=\!\! 
	\begin{bmatrix}
		\bzeta^T_{{\mathbf O}_{k \textendash N+1}} \!&\!
		\cdots \!&\! 
		\bzeta^T_{{\mathbf O}_{k \textendash 1}}
	\end{bmatrix}^T,
	\bzeta_{\mathbf O_{i}}
	\!\!=\!\! 
	\begin{bmatrix}
		^{\mathbf G} \mathbf p^T _{\mathbf O_i} \!&\! 
		^{\mathbf G} _{\mathbf O_i} \bar{\mathbf q}^T 
	\end{bmatrix}^T
\end{align}
for $i=k-N+1,\,\cdots,\,k-1$. 

Once a new image is received, keyframe selection, pose integration (see Sec.~\ref{sec: 6d mani inte}), and image processing (with FAST feature~\cite{rosten2008faster} and FREAK~\cite{alahi2012freak} descriptor) are conducted in sequential order. Since they are standard steps for sliding-window pose estimation~\cite{zhang2019large,qin2017vins}, we omit the details here.
Subsequently, we minimize the following cost function to refine our state:
\color{black}
\begin{align}
	\label{eq:total cost}
	&\mathcal{C}_{k+1}
	(\mathbf y_k, \bst_{\mathbf {k+1}}) = 
	2\, \boldsymbol \eta_{k}^T {} ( \mathbf y_k - \hat{\mathbf y}_k )
	+ || \mathbf y_k - \hat{\mathbf y}_k ||_{\boldsymbol \Sigma_{k}} \notag \\  &+
	\sum _{i,j \in \mathbf S_{i,j}} \gamma_{i,j} 
	+
	\sum _{i = k-N+1}^{k+1}  \psi_i +
	\beta(\bst_k, \bst_{k+1})
\end{align}
\color{black}
where $ \boldsymbol \eta_{k}$ and $\boldsymbol \Sigma_{k}$ are estimated prior information vector and matrix, which can be obtained from marginalizing states and measurements to probabilistically maintain bounded computational cost. For marginalization details please refer to \cite{forster2017manifold,qin2017vins}.
Furthermore, in the above equation, $||\mathbf a||_{\boldsymbol \Sigma}$ is computed by $\mathbf a^T {\boldsymbol \Sigma} \mathbf a$.
Besides,
$ \rev{\gamma_{i,j}}$ is the computed camera reprojection residuals;
$ \rev{\beta}$ \rev{corresponds to the prediction cost} by odometer measurements between time $t_k$ and $t_{k+1}$, as described in Sec.~\ref{sec: manifold formal int}; and $ \rev{\psi_i}$ is the residual corresponding to the motion manifold which will be illustrated in the Sec.~\ref{sec:mani}.%

%

Specifically, the image induced cost function is:
\begin{align}
	 \gamma_{i,j} =
	{
		\left\|
		\mathbf z_{ij} - h \big{(} \bzeta_{\mathbf O_i}, \mathbf f_j \big{)}
		\right\|
	}_{\boldsymbol{\Sigma}_C}
\end{align}
where $\mathbf z_{ij}$ represents 
camera measurement corresponding to the pose $i$ and visual landmark $\mathbf{f}_j$. Moreover, $\mathbf S_{i,j}$ represents the set of pairs between keyframes and observed features.
$\boldsymbol{\Sigma}_C$ is the measurement information matrix,
and the function $h( \cdot )$ is the model of a calibrated perspective camera~\cite{li2014online}. 
To deal with landmark state $\mathbf{f}_j$, we choose to perform multiple-view triangulation, optimization, and then stochastic marginalization~\cite{li2013high, mourikis2007} to keep a relatively low computational cost.

Since IMU is also a widely used low-cost sensor, we can also use measurements from an additional IMU to further improve the estimation accuracy.
Specifically, when an IMU is used in combination with wheel odometer and a monocular camera, the state vector of our algorithm (see Eq.~\eqref{eq:state}) becomes:
\begin{align}
	\label{eq:state_imu}
	\mathbf y^\prime_k 
	= 
	\begin{bmatrix}
		{\mathbf O_k}^T &
		\bst_k^T &
		\rev{\boldsymbol \mu}_k^T &
	\end{bmatrix}^T\!,
	\rev{\boldsymbol \mu}_k = 
	\begin{bmatrix}
		^{\mathbf G} \mathbf v _{\mathbf I_k}^T &
		\mathbf b_{\mathbf a_k}^T &
		\mathbf b_{\mathbf g_k}^T &
	\end{bmatrix}^T
\end{align}
where $^{\mathbf G} \mathbf v _{\mathbf I_k}$ is IMU's velocity expressed in global frame, and $\mathbf b_{\mathbf a_k}$ and $\mathbf b_{\mathbf g_k}$ are accelerometer and gyroscope biases respectively.
Once a new image is received and a new keyframe is determined, we optimize the IMU-assisted cost function:
\begin{align}
	\mathcal{C}^\prime_{k+1}
	(\mathbf y^\prime_k, \bst_{\mathbf {k+1}},
	\rev{\boldsymbol \mu}_{k+1}) &= 
	\mathcal{C}_{k+1}
	(\mathbf y_k, \bst_{{k+1}})
	\notag \\
	&+\kappa (\bzeta_{\mathbf O_{k}},  \bzeta_{\mathbf O_{k+1}}, \rev{\boldsymbol \mu}_{k}, \rev{\boldsymbol \mu}_{k+1})
	\label{eq:total cost1}
\end{align}
Compared to Eq.~\eqref{eq:total cost} when IMU is not used, 
Eq.~\eqref{eq:total cost1} only introduce an additional IMU cost $\kappa(\cdot)$. Since IMU cost functions are mature and described in a variety of existing literatures~\cite{li2013high,leutenegger2015keyframe,forster2017manifold}, we here omit the discussion on detailed formulation of $\kappa (\cdot)$.
Overall, the work flow of the system can be summarized in Alg.~\ref{alg:Framework}.

\subsection{Manifold Constraints}
\label{sec:mani}
In the system, in addition to the \rev{manifold based prediction} cost $\beta$, we also introduce additional cost term $\boldsymbol  \psi$  into Eq.~\eqref{eq:total cost}, which provides explicitly kinematic constraints for multiple poses in the sliding window based on the motion manifold~\cite{zhang2019large}. 

Specifically, if a local motion manifold is perfectly parametrically characterized at around $^{\mathbf G} \mathbf p _{\mathbf O_k}$,
the following equation holds:
\begin{align}
\label{eq:manifold11}
\mathcal{M}(^{\mathbf G} \mathbf p _{\mathbf O_i}) = 0,\text{for}\,\text{all}\,i \,\, \text{that}\,\,\, ||{} ^{\mathbf G} \mathbf p _{\mathbf O_k} - {}
^{\mathbf G} \mathbf p _{\mathbf O_i}|| < \epsilon
\end{align} where $\epsilon$ is a distance \rev{threshold}, representing the region size that the local manifold spans.
Since modeling errors are inevitable for manifold representation, 
we model Eq.~\eqref{eq:manifold11} as a stochastic constraint. 
Specifically, we define \rev{the manifold constraint residual as} $\boldsymbol \psi_i = r_{p_i} + r_{q_i}$, with 
\begin{align}
	\label{eq:pq}
	r_{p_i} = 
	\frac{1}{\boldsymbol \sigma^2_p}
	\left( \mathcal{M} \big( {^G \mathbf p_{\mathbf O_i}} \big) \right)^2
\end{align}
and ${\boldsymbol \sigma^2_p}$ is the corresponding noise variance.
Additionally, we apply an orientation cost which origins from the fact that the orientation of the ground robot is affected by the manifold profile, as shown in Fig.~\eqref{fig:robot} \rev{and Eq.~\eqref{eq:int1}}:
\begin{align}
	\label{eq:fq}
	r_{q_i} = 
	\left\| 
	\lfloor {^{\mathbf G}_{\mathbf O_i} \mathbf R} \cdot {\mathbf e_3}
	\rfloor_{12} \cdot
	\frac{\partial \,\, \mathcal{M}}{\partial \,\, \mathbf p} \bigg{|}_{\mathbf p = {^G \mathbf p_{\mathbf O_i}}}
	\right\|_{\boldsymbol{\Omega}_q}
\end{align}
where $\boldsymbol{\Omega}_q$ is the measurement noise information matrix, \rev{and $\nabla \mathcal{M}
	\left( {^\bG \mathbf p _{\bO_i}} \right) =  \frac{\partial \,\, \mathcal{M}}{\partial \,\, \mathbf p} \bigg{|}_{\mathbf p = {^G \mathbf p_{\mathbf O_i}}}$}.

We point out that, the manifold parameters used in propagation can be considered as {\em implicit} constraints, providing frame-to-frame motion information. \rev{On the other hand, $\psi_i$ can be treated as {\em explicit} multi-frame constraints for state optimization. Theoretically, the implicit constraints compute Jacobians with respect to manifold parameters, which allow for probabilistic updates. However, we find that by imposing the additional explicit constraints, the corresponding states can be better estimated. Thus, we incorporate $\psi_i$ into the cost functions in estimator formulation.}

Furthermore, we also take the operations in Eq.~\eqref{eq:per axis m} to make the manifold parameters adapted to the changes of the manifold, as the method described in Sec.~\ref{sec:re-param} to re-parametrize the manifold parameters at {\em each} step after the state marginalization. This step is applied to both the estimated manifold parameters and the corresponding prior information matrix. 
We also note that in Eq.~\eqref{eq:manifold} we choose to fix the coefficient of $z$ to be \rev{a constant number of} 1, which means that our manifold representation is {\em not} for generic motion \rev{in 3D space}. However, this perfectly fits our applications, since most ground robots can {\em not} climb vertical walls.

The last algorithmic component that needs to be described is manifold initialization. In this work, we adopt a simple solution for the initialization of the manifold parameters. Specifically, in the odometer-camera system, the initial global frame is defined to be aligned with the initial odometer frame. The initial motion manifold is defined to be the tangent plane corresponding to the odometer frame, i.e., the planar surface defined by the x and y axes of the odometer frame. On the other hand, we assume zero values for the second-order terms in the initial guess of the manifold parameter vector, and allocate initial information matrix to capture the corresponding uncertainties.

\section{EXPERIMENTS}
\label{sec:exp} 
\renewcommand{\arraystretch}{1.0} 

\begin{figure}[tbp]
	\subfigure{
		\label{fig:sim:a}
		\includegraphics[width=\columnwidth]
		{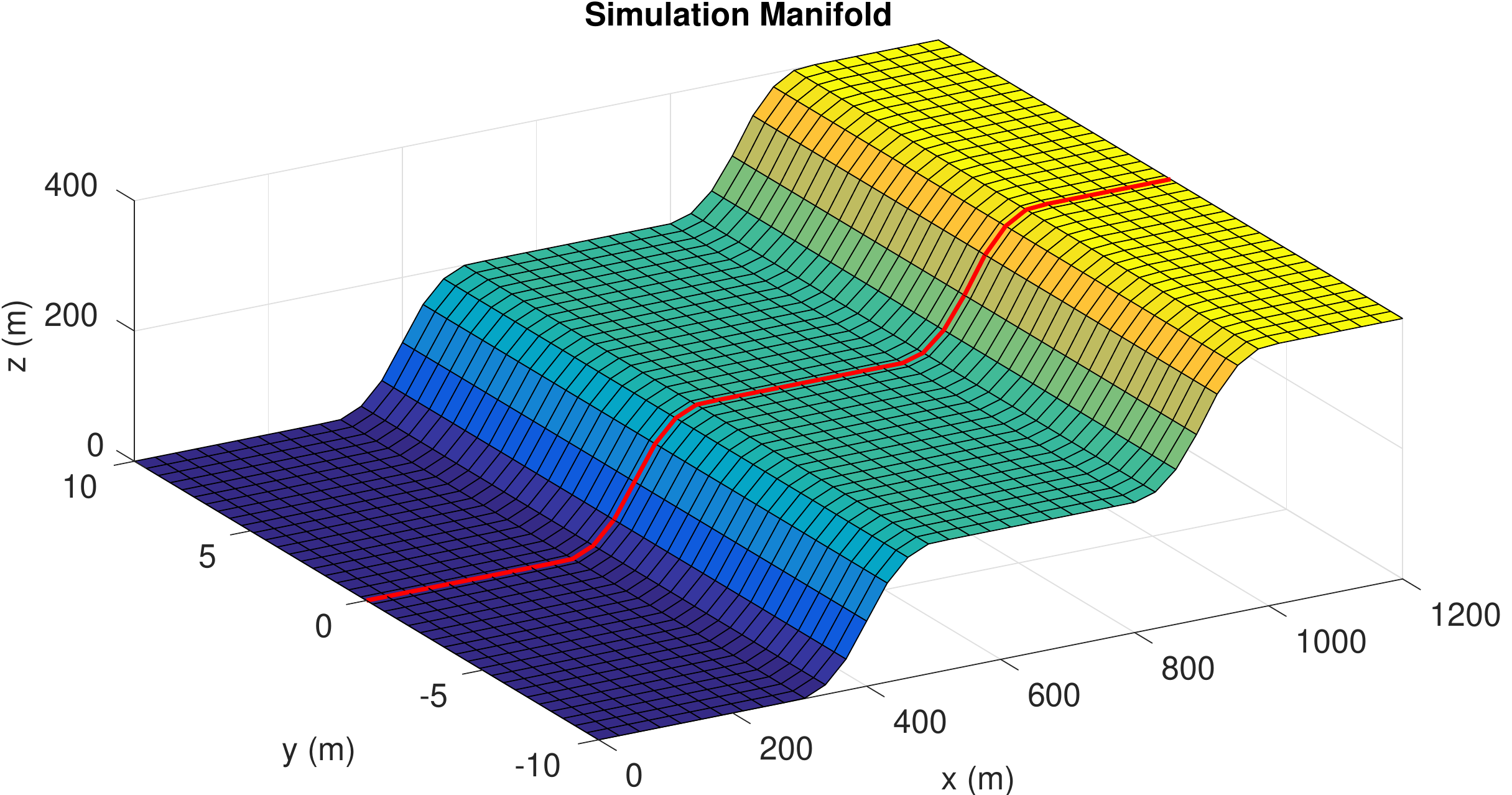}
	}    
	\subfigure{
		\label{fig:sim:b}
		\includegraphics[width=\columnwidth]
		{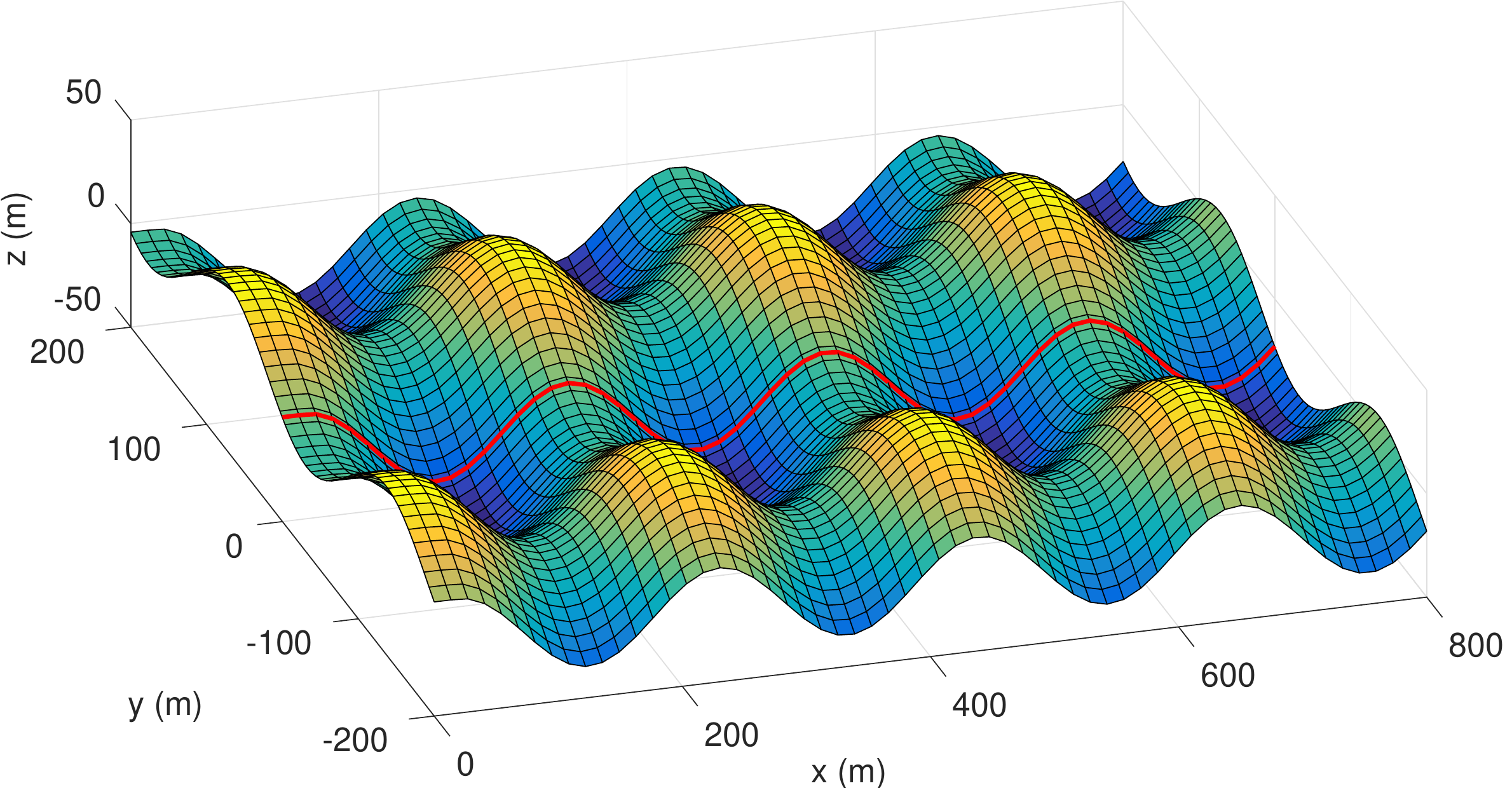}
	}
	\caption{Motion manifold used in simulation tests with red solid line trajectory. Top: Motion manifold constructed by piecewise quadratic polynomials. Bottom: Motion manifold constructed by sinusoidal and cosusoidal functions. 
           }\label{fig:sim-mani}
\end{figure}


In this section,  we show results from both simulation tests and real-world experiments 
to demonstrate the performance of the proposed methods.

\subsection{Simulation Tests}\label{sec:simu exp}
\begin{figure}[tbp]
	\subfigure{
		\includegraphics[width=\columnwidth]
		{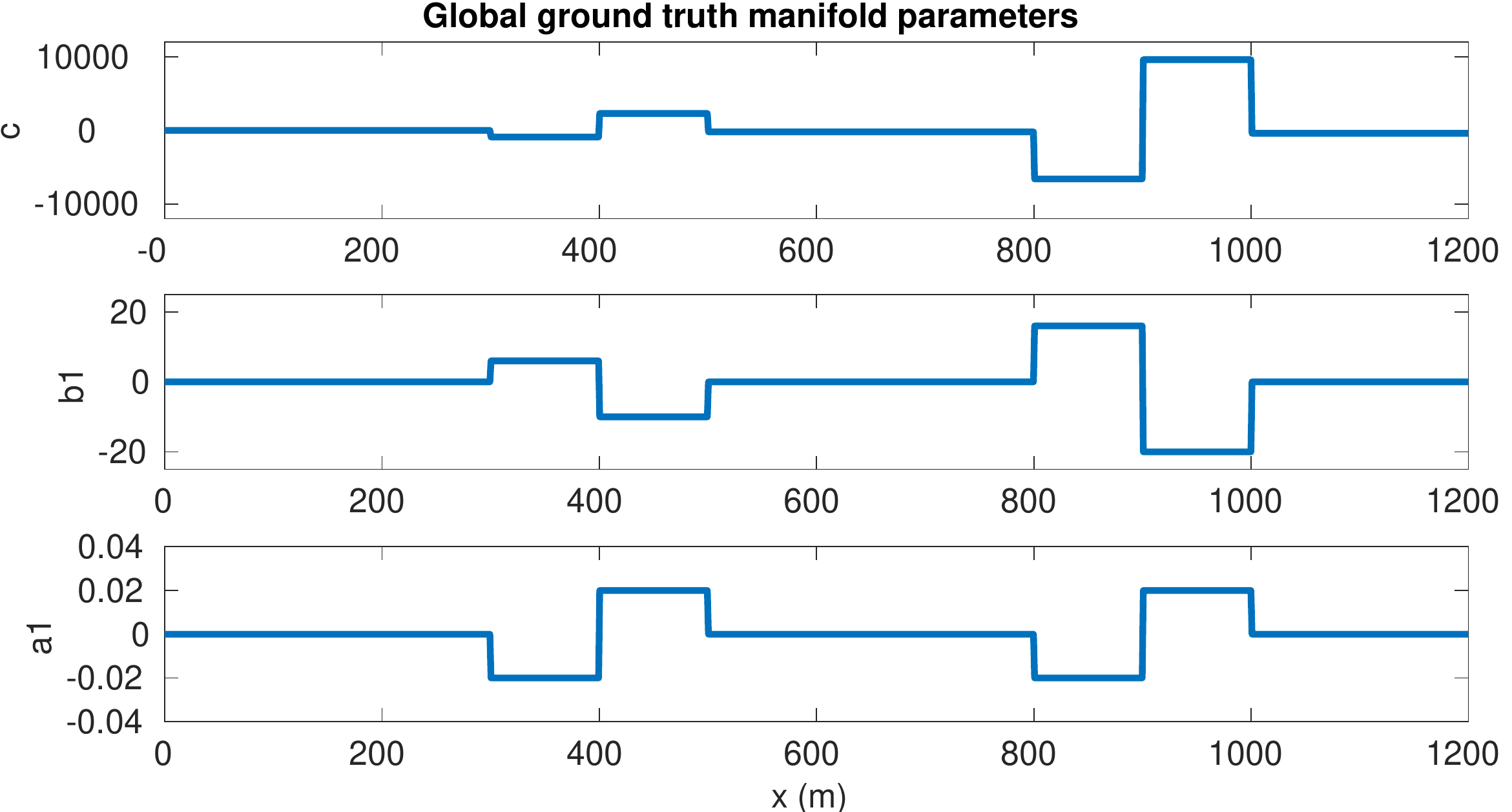}
	}
	\caption{Ground truth global manifold parameters used for generating simulation trajectory shown in the upper plot of Fig.~\ref{fig:sim-mani}. Only non-zero elements are plotted, while $b_2$, $a_2$, and $a_3$ are constantly zero in this piecewise polynomial test. }\label{fig:sim-mani-p}
\end{figure}

In the simulation tests, we assumed a ground robot moving on a manifold, whose gradient vectors are continuous in 
$\mathbb{R}^3$. 
Specifically, we generated two types of manifold profile for simulations, 
shown in Fig.~\ref{fig:sim-mani}. The first type of motion manifold (top one in Fig.~\ref{fig:sim-mani}) is generated by the piecewise polynomial function, and the corresponding manifold parameters are plotted in~Fig.~\ref{fig:sim-mani-p}. 
Although the manifold visualized in Fig.~\ref{fig:sim-mani} seems smooth, we note that the corresponding global parameters change significantly (see~Fig.~\ref{fig:sim-mani-p}).
By simulating motion manifold in this setup, the overall performance of the proposed algorithm and the capability of handling changes of motion manifold can be evaluated. 
The second type (bottom one in Fig.~\ref{fig:sim-mani}) of manifold is generated by planar and sinusoidal functions. This is to simulate the fact that in real-world scenarios `quadratic' equations are basically `local' approximations, which in general cannot perfectly characterize all road conditions. 
With this simulated environment, we are able to test the performance of the proposed method for approximating road conditions with locally quadratic equations.

In the tests, simulated odometer, IMU, and camera measurements were generated at 
$100$~Hz, $100$~Hz, and $10$~Hz, respectively. Extrinsic parameters between sensors
are assumed to be rigidly fixed, and intrinsic parameters of sensors are assumed to be perfectly calibrated. To obtain camera measurements efficiently, we directly generated feature measurements for each image, with simulated 3D feature positions and known data association. We generated $400$ feature points per image on average, and the averaged feature
track length is about $5.1$ frames.
\rev{During the simulations, the robot moves forward with a constant local velocity at $3.5m/s$.}
%
To make the simulation tests more realistic, the sensor noise distribution used to generate measurements are characterized by the real sensors used in our experiments (see Sec.~\ref{sec:exp2}).  Specifically, we assume zero-mean Gaussian distribution for all sensor noises, and the corresponding standard deviations are as follow: $0.8$ pixel per feature observation for image, $9\cdot10^{-4} rad/s^{0.5}$ for gyroscope measurement\footnote{Note, this is continuous-time IMU noise, see~\cite{trawny2005indirect} for details.}, $1\cdot10^{-4}rad/s^{1.5}$ for gyroscope bias, $1\cdot10^{-2}m/s^{1.5}$ for accelerometer measurement, $1\cdot10^{-4}m/s^{2.5}$ for accelerometer biases, and finally $3\%$ percentage error for wheel encoder readings with $9.8cm$ wheel radius and $38cm$ left-to-right wheel baseline.

\begin{table}[t]  
	\centering  
	\begin{threeparttable}  
		\caption{Pose integration errors in \rev{simulated piecewise polynomials scenario}. Three methods are compared: IMU-based integration (IMU method), traditional odometer integration that relies on planar surface assumptions (Trad. Odom.), and the proposed manifold based integration (Mani. Odom.).}  
		
		\label{table: pose integration}
		\begin{tabular}{l p{0.4cm}<{\centering}
				p{0.4cm}<{\centering}
				p{0.4cm}<{\centering}
				p{0.4cm}<{\centering} 
				p{0.4cm}<{\centering} 
			}  
			\toprule  
			\multirow{2}{*}{} \bf{Integ. Time} &  
			\multicolumn{1}{c}{ 
				{$0.1$~sec}}&\multicolumn{1}{c}{
				{$1.0$~sec}}&\multicolumn{1}{c}{
				{$3.0$~sec}}&\multicolumn{1}{c}{
				{$5.0$~sec}}&\multicolumn{1}{c}{    
				{$10.0$~sec}}\cr  
			
			\midrule
			
			\multirow{2}{*}{} \bf{Traj. Lengths} &  
			\multicolumn{1}{c}{ 
				$0.35$~m}&\multicolumn{1}{c}{
				$3.5$~m}&\multicolumn{1}{c}{
				$10.5$~m}&\multicolumn{1}{c}{
				$17.5$~m}&\multicolumn{1}{c}{    
				$35.0$~m}\cr  
			
			\midrule
			
			\multicolumn{1}{l}{\bf{IMU method}} \cr
			\quad pos. err. (m)    &0.0027 & 0.0093  & 0.0544 &  0.1695&  0.9329\\
			\quad rot. err. (deg.) & 0.0249 & 0.0794 & 0.1450 &  0.1975 &  0.2863 \\
			
			\midrule
			\multicolumn{1}{l}{\bf{Trad. Odom.}} \cr
			\quad pos. err. (m)  &  0.0028  & 0.0260 &0.1899  & 0.5116  &  2.1637  \\
			\quad rot. err. (deg.) & 0.0701 & 0.6856 & 1.9715  &  3.2468  & 6.8433\\
			
			\midrule
			\multicolumn{1}{l}{\bf{Mani. Odom.}} \cr
			\quad pos. err. (m)    &0.0026 & 0.0086  & 0.0225 &  0.0372 &  0.0688 \\
			\quad rot. err. (deg.)&  0.0205  & 0.0646& 0.1221 &  0.1530 &  0.1621 \\
			
			\bottomrule  
		\end{tabular}  
	\end{threeparttable}  
\end{table} 

\definecolor{Gray1}{gray}{0.85}
\definecolor{Gray2}{gray}{0.65}
\newcolumntype{G}{>{\columncolor{Gray1}}c}
\newcolumntype{g}{>{\columncolor{Gray2}}c}
\begin{table*}[tp]  
	\centering  
	\begin{threeparttable}  
		\caption{Simulation results: pose estimation errors for different methods.}  
		
		\label{table: sim loc}
		\begin{tabular}{l 
				G G G G G g g g g
			} 
			\toprule  
			\multirow{2}{*}{} 
			\bf{Algorithm} 
			& \multicolumn{1}{c} {\bf Mn }
			& \multicolumn{1}{c} {\bf M0th}
			& \multicolumn{1}{c} {\bf M1st}
			& \multicolumn{1}{c} {\bf M2nd}
			& \multicolumn{1}{c} {\bf M2nd+Re}
			& \multicolumn{1}{c} {\bf MSCKF\cite{li2013optimization}}
			& \multicolumn{1}{c} {\bf VINS-W\cite{wu2017vins}}
			& \multicolumn{1}{c} {\bf AM+IMU\cite{zhang2019large}}            
			& \multicolumn{1}{c} {\bf{M2nd+Re+IMU}}
			\cr  
			
			\midrule
			
			\bf{w. IMU} 
			& \multicolumn{1}{c} {\bf No }
			& \multicolumn{1}{c} {\bf No}
			& \multicolumn{1}{c} {\bf No}
			& \multicolumn{1}{c} {\bf No}
			& \multicolumn{1}{c} {\bf No}
			& \multicolumn{1}{c} {\bf Yes}
			& \multicolumn{1}{c} {\bf Yes}
			& \multicolumn{1}{c} {\bf Yes}            
			& \multicolumn{1}{c} {\bf Yes}
			\cr  
			
			\midrule
			
			\multicolumn{1}{c}{\bf{Piecewise Env.}} \cr
			{\bf{Pos. err. (m)}} & 
			8.073  & 
			39.031 & 
			18.926 &  
			11.321 &  
			1.043 & 
			16.400 & 
			5.622  &
			0.673  &
			\bf{0.323} \\
			\bf{Rot. err. (deg)} & 
			1.839 & 
			7.501 & 
			3.256 & 
			1.719 & 
			0.246  & 
			0.115&  
			0.323 &  
			0.109& 
			\bf{0.099} \\        
			
			\midrule
			\multicolumn{1}{c}{\bf{Sinusoidal Env.}} \cr
			{\bf{Pos. err. (m)}} & 
			56.391 & 
			291.032 & 
			176.329 &              
			78.335 & 
			0.663 & 
			18.695 &  
			13.097  &  
			0.422& 
			\bf{0.389}\\
			{\bf{Rot. err. (deg)}} &
			1.083 & 
			89.237 & 
			45.093 &              
			25.665 & 
			0.066 & 
			0.371 &  
			0.117 &  
			0.090 & 
			\bf{0.057} \\
			\bottomrule  
		\end{tabular}  
	\end{threeparttable} 
\end{table*}

\subsubsection{Pose Integration Tests}
The first test is to demonstrate the \rev{theoretical performance} of the 6-DoF pose integration by using the proposed method, \rev{with known initial conditions}.  Specifically, we compared the proposed manifold-based integration methods
against the widely used IMU-based pose integration~\cite{li2013high,qin2017vins} and \rev{traditional 
odometer-based} integration~\cite{censi2013simultaneous,wu2017vins} which relies on planar surface assumption. 
These tests are designed for forward pose integrations without constraints from visual sensors to demonstrate the integration performance.
\rev{The pose integration was performed for different short periods from $0.1$ to $10$ seconds (corresponding to trajectory lengths from $0.35$ to $35$ meters), in regions where non-planar slopes exist in the top plot of Fig.~\ref{fig:sim-mani}.}
\rev{Additionally, since we focused on pose integration in this test, the estimation systems were assumed to be free of errors in initial conditions and only subject to measurement errors}. \rev{We assumed:} 1) the motion manifold is precisely known for the proposed method, 2) correspondingly, initial roll and pitch angles of the IMU are free of errors in the IMU method, and 3) initial velocity is known precisely for all methods. In real applications, all those variables will be estimated online and subject to estimation errors, and in the next section we will show the detailed performance of integrated systems.
%
\rev{Table~\ref{table: pose integration} shows the averaged position and orientation errors for different methods over 300 Monte-Carlo runs.}
%
Those results show that by explicitly incorporating manifold equations into the odometer integration process, the accuracy can be improved \rev{compared to the traditional planar surface assumption}.
\rev{On the other hand, the accuracy of short-term IMU integration is comparable with our proposed method, while for long-term integration tasks, the proposed method performs certainly better.}

\subsubsection{Pose Estimation Tests}
\label{exp:sim loc}
The second test is to demonstrate the pose estimation performance of the proposed method, \rev{when integrated into a sliding-window estimator with measurements from a monocular camera, wheel encoders, and optionally an IMU}. Specifically, we implemented nine algorithms in this test.
	a) {\bf Mn }: the traditional camera and odometer fusion approach without manifold modeling~\cite{jesus2012combining}. Specifically, in traditional methods, linear and rotational velocities measured by wheel encoders can be used directly for an update. Alternatively, those velocity terms can be integrated to compute 3D relative constraints and allocate uncertainty terms in all 6D directions. In this work, we adopt the second implementation;
	b) {\bf M0th}: the proposed method with zeroth order manifold model to fuse odometer and camera measurements;
	c) {\bf M1st}: the proposed method with first order manifold model;
	d) {\bf M2nd}: the proposed method with second order manifold model;
	e) {\bf M2nd+Re}: the proposed method with second order manifold model and utilize manifold re-parameterization. This is our theoretically preferred method when IMU is not included;
	f) {\bf MSCKF\cite{li2013optimization}}: standard visual-inertial odometry;
	g) {\bf VINS-W~\cite{wu2017vins}}: the method representing the current state-of-the-art technique for fusing measurements from a monocular camera, the wheel odometer, and an IMU;
	h) {\bf AM+IMU}~\cite{zhang2019large}: our previous work of using wheel odometer, a monocular camera, and an IMU for pose estimation, by approximating the manifold based integration process. Specifically, compared to the proposed method in this work, our previous work \cite{zhang2019large} utilizes planar surface based odometry integration for predicting estimated poses, assumes uncertainties in all 6D directions, and does not introduce the manifold re-parameterization process;
	and finally 
	i) {\bf{M2nd+Re+IMU}}: proposed second order manifold method with re-parameterization using odometer, camera and IMU measurements.
For all different methods, we compute root-mean-squared-errors (RMSEs) for both 3D position estimates and 3D orientation estimates, which are shown in Table~\ref{table: sim loc}. 
Note that the five camera-odometer fusion approaches: {\bf Mn }, {\bf M0th}, {\bf M1st}, {\bf M2nd} and {\bf M2nd+Re} are colored with light gray in Table~\ref{table: sim loc} (as well as Table~\ref{table:kasit_highway} and~\ref{table:kasit_urban} later), while those IMU-integrated approaches:
	{\bf MSCKF\cite{li2013optimization}},
	{\bf VINS-W~\cite{wu2017vins}},  {\bf AM+IMU~\cite{zhang2019large}}
	and {\bf{M2nd+Re+IMU}} are in dark gray.

A couple of conclusions can be made from the results. Firstly, 
our proposed method is able to obtain \rev{accurate} estimation results even without using IMUs. This is not achievable by other competing methods. 
In fact, the accuracy of the proposed method is even better than a couple of competing methods that employ an IMU sensor.
Secondly, when an IMU is also used in the system, the proposed method yields the best performance among algorithms tested. This result indicates that the proposed method is the most preferred one for conducting high-accuracy estimation.
Thirdly, we also noticed that when manifold re-parameterization is not used, the errors increase significantly. This validates our theoretical analysis in Sec.~\ref{sec:re-param} that the re-parameterization is a necessary step to ensure manifold propagation errors can be characterized statistically.
In addition, we find that if the motion manifold is not 
properly parameterized, the estimation precision will be significantly reduced. Specifically, both zeroth order and first order assumptions do not allow for 6D motion and inevitably lead to extra inaccuracy. On the other hand, the proposed {\bf M2nd+Re} is able to model 6D motion via manifold representation, and yield good pose estimates.
Finally, it is also important to note that the proposed methods outperform competing methods also under the sinusoidal manifold. This is due to the fact that our quadratic representation of motion manifold is effectively locally within each sliding window, and thus it is able to approximate complex real-scenarios without having large modeling errors.
We also note that, when used in the \rev{piecewise planar} simulation environment, the estimation errors of VIO are large. This is due to its theoretical drawback of having extra un-observable space under constant-velocity straight-line  
motion~\cite{li2013high}. Since this type of motion is common for the ground robots, adding the odometer sensor for pose estimation becomes important.

\begin{figure*}[t] 
	\begin{minipage}{.33\linewidth}
		\hspace*{-0.1em}
		\subfigure{
			\includegraphics[width=\columnwidth]
			{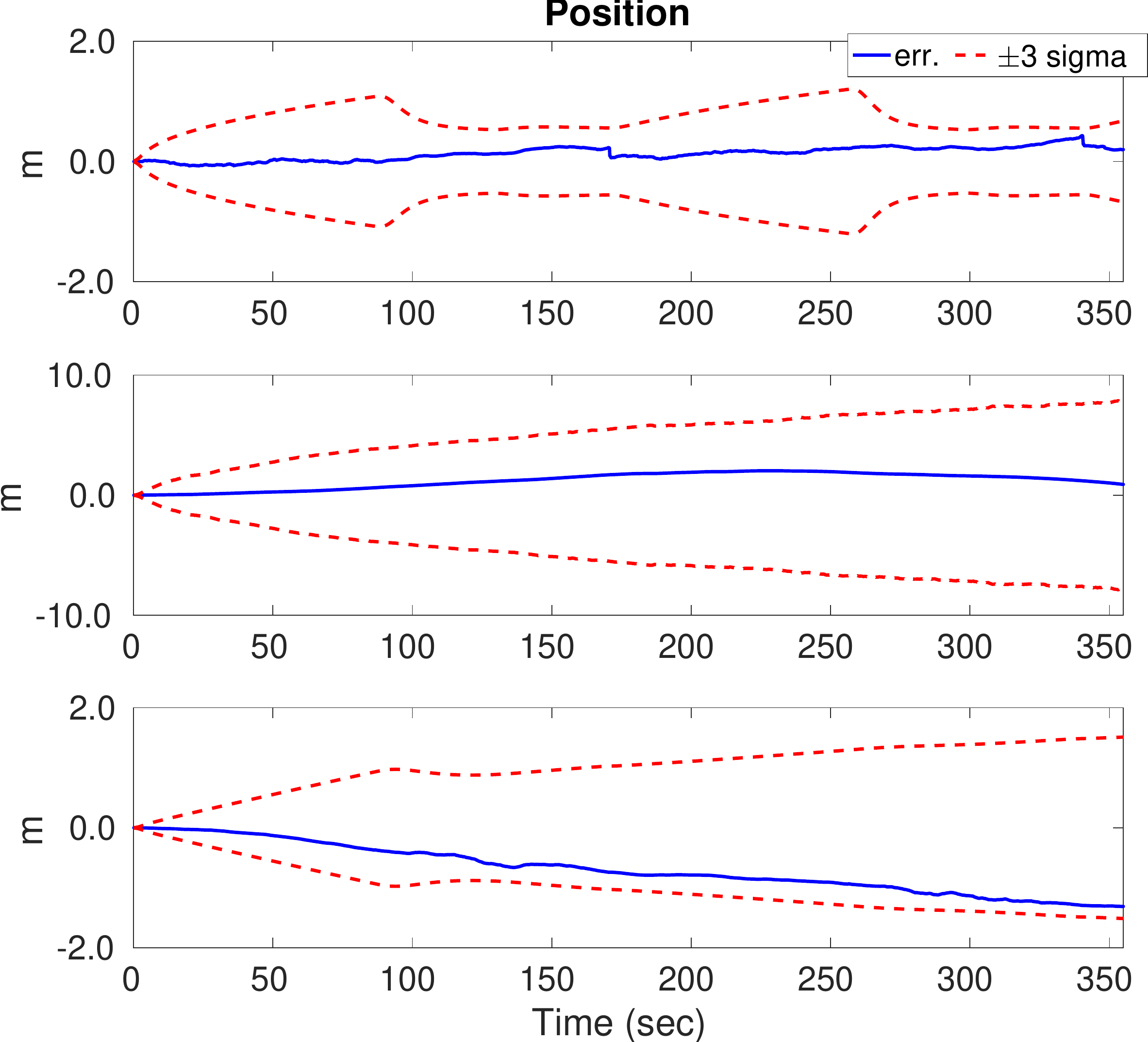}
		}
	\end{minipage}
	\begin{minipage}{.33\linewidth}
		\hspace*{-0.1em}
		\subfigure{
			\includegraphics[width=\columnwidth]
			{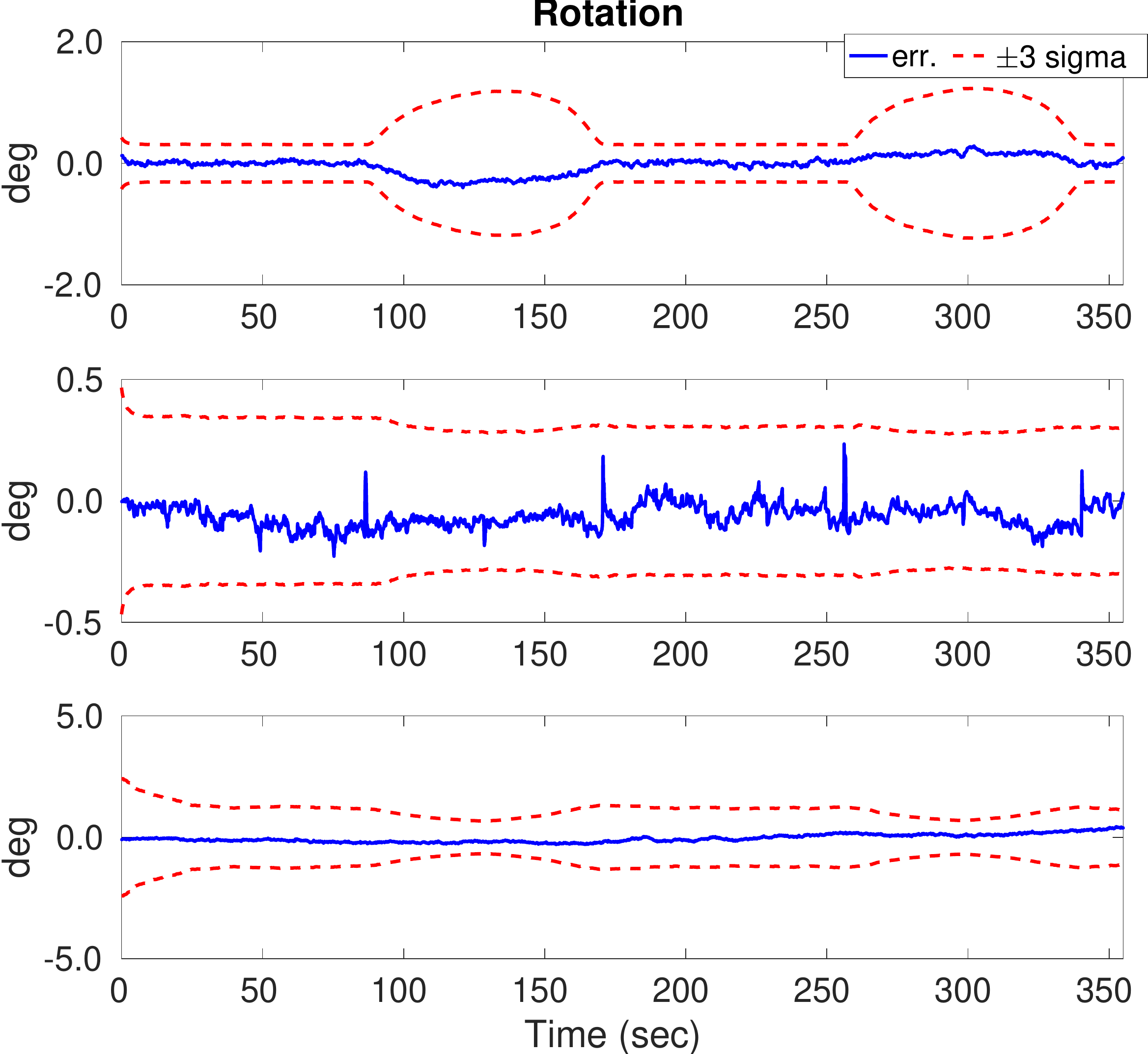}
		}
	\end{minipage}
	\begin{minipage}{.33\linewidth}
		\hspace*{-0.1em}
		\subfigure{
			\includegraphics[width=\columnwidth]
			{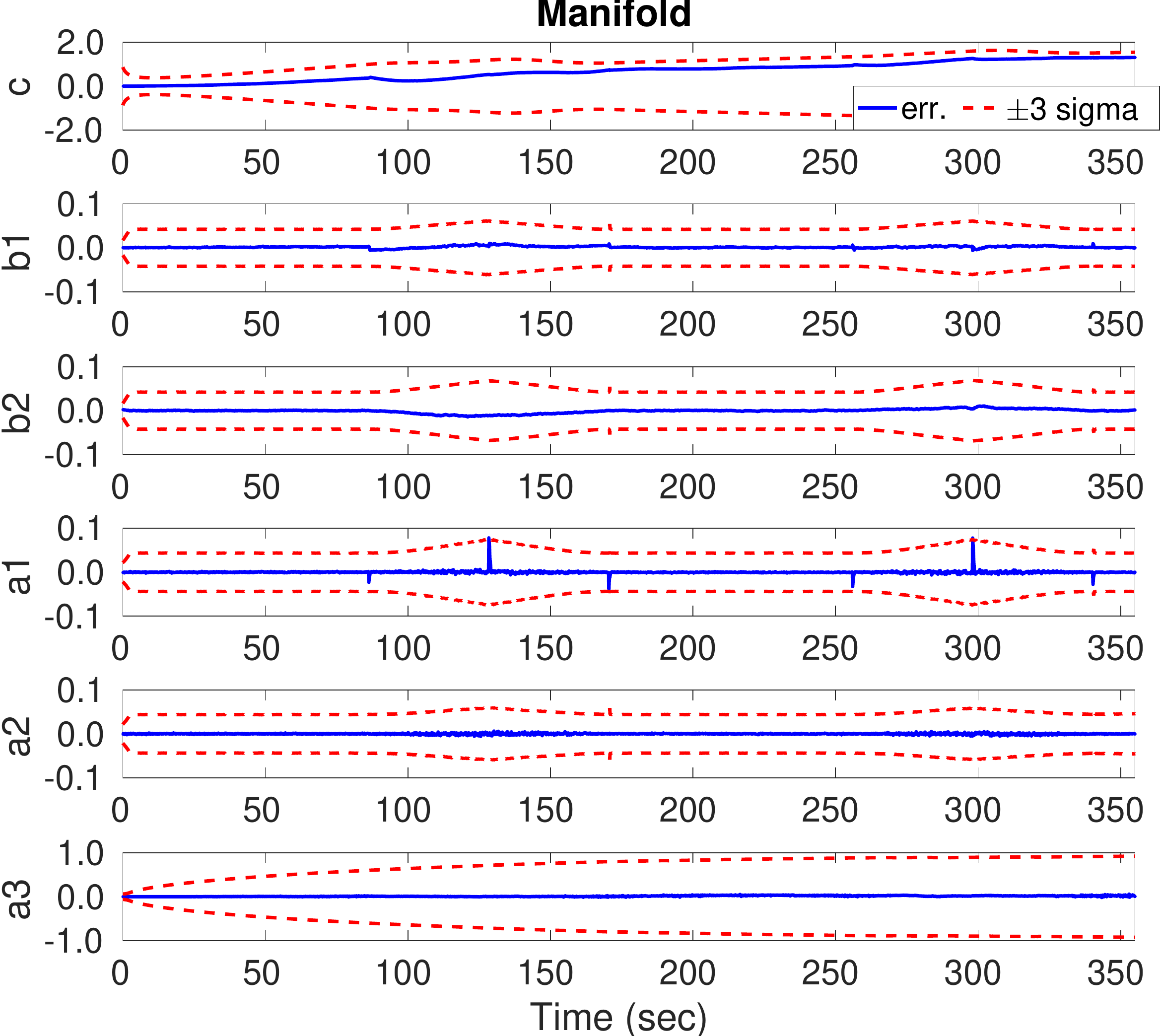}
		}
	\end{minipage}
	\caption{For the simulated test on the upper plot of Fig.~\ref{fig:sim-mani}, state estimation errors and the corresponding estimated $\pm3 \sigma$ bounds \rev{for {\bf M2nd+Re}}. The left subplots corresponds to 3D position (x, y, z from \rev{top} to bottom), the middle ones to 3D orientation (pith, roll, yaw from \rev{top} to bottom), and the right ones to the motion manifold parameters.
	} \label{fig:sim_loc}
\end{figure*}

Additionally, we also plotted the estimation errors and the corresponding \rev{$3\sigma$ uncertainty bounds} for the 3D position, 3D orientation, and manifold parameters of the proposed method \rev{\bf M2nd+Re} in Fig.~\ref{fig:sim_loc}, which are from a representative run in the upper plot of Fig.~\ref{fig:sim-mani}. The most important observation is that the manifold parameters can be accurately estimated, and the estimation errors are well characterized by their uncertainties bounds. Additionally, we also noticed that there are two `circle' style curves in the uncertainty \rev{bounds} of the first axis of orientation error. 
In fact, those are corresponding to the regions when the ground robot was climbing \rev{by continuously varying orientation in pitch}. 
\rev{Consequently, manifold propagation and re-parameterization compute the corresponding Jacobians and lead to the changes happening in the uncertainty bounds.}
Finally, we also point out that both positional and orientational estimates are well characterized by the corresponding uncertainty \rev{bounds} in the proposed methods.

\subsection{Real-World Experiments}
\subsubsection{Testing Platforms and Environments}
To evaluate the performance of our proposed approach, 
we conducted experiments by using datasets from both  the ones that are publicly available as well as our customized sensor platform. 

We first carried out tests using dataset sequences from the publicly available KAIST dataset~\cite{jeong2019complex}, which was collected by ground vehicles in different cities in South Korea. The sensors used in our experiment were the left camera of the equipped Pointgrey Flea3 stereo system, the Xsens MTi-300 IMU, and the RLS LM13 wheel encoders. The measurements recorded by those sensors were $10$Hz, $100$Hz, and $100$Hz, respectively. 
On the other hand, our customized sensor platform consists of a stereo camera system with ON AR0144 imaging sensors, a Bosch BMI 088 IMU, and wheel encoders.
For our proposed algorithms, only the left eye of the stereo camera system was used.
During our data collection, images were recorded at $10$Hz with $640\times400$-pixels resolution, IMU measurements
were at $200$Hz, and wheel odometer measurements were at $100$Hz. 
Our datasets were all collected on the Yuquan campus of 
Zhejiang University, Hangzhou, China. This campus is located next to the West Lake in Hangzhou, embraced by mountains on three sides. Therefore, the environments in our datasets are \rev{complex}, including frequent slope changes and terrain condition changes. 

In terms of performance evaluation, we used both the final positional drift and RMSE to represent the estimation accuracy. For the dataset collected by our platform, due to lack of full-trajectory ground truth, we dedicatedly started and ended our data collection process at exactly the same position, which enables computing the final drift. On the other hand, ground truth poses for the KAIST dataset~\cite{jeong2019complex} are available, and we used that for qualitative performance evaluation by computing RMSE.

\begin{figure}[htbp] 
	\centering 
	\subfigure{ 
		\includegraphics[width=\columnwidth]{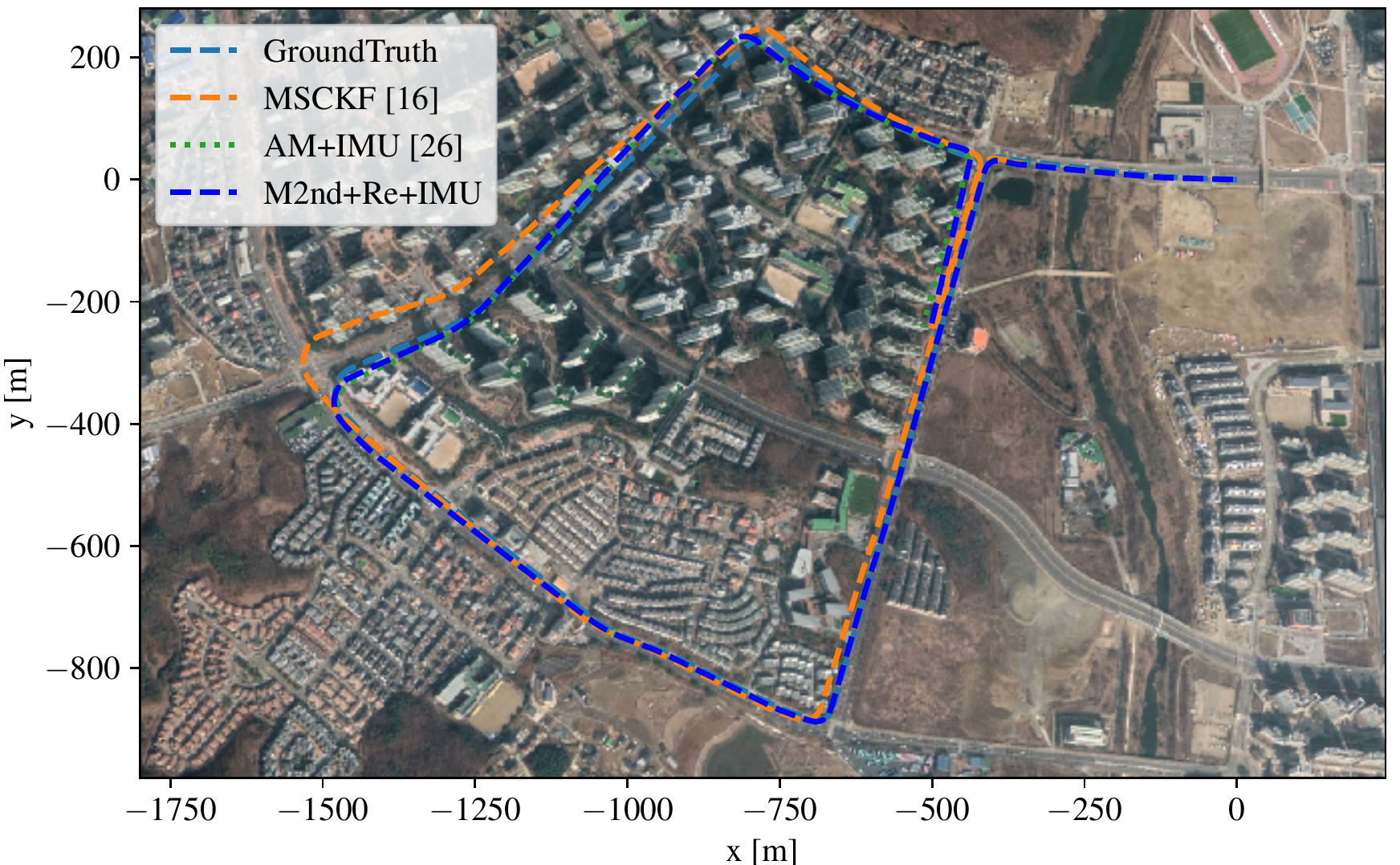} 
	}\hspace*{-0.0em}
	
	\subfigure{ 
		\includegraphics[width=\columnwidth]{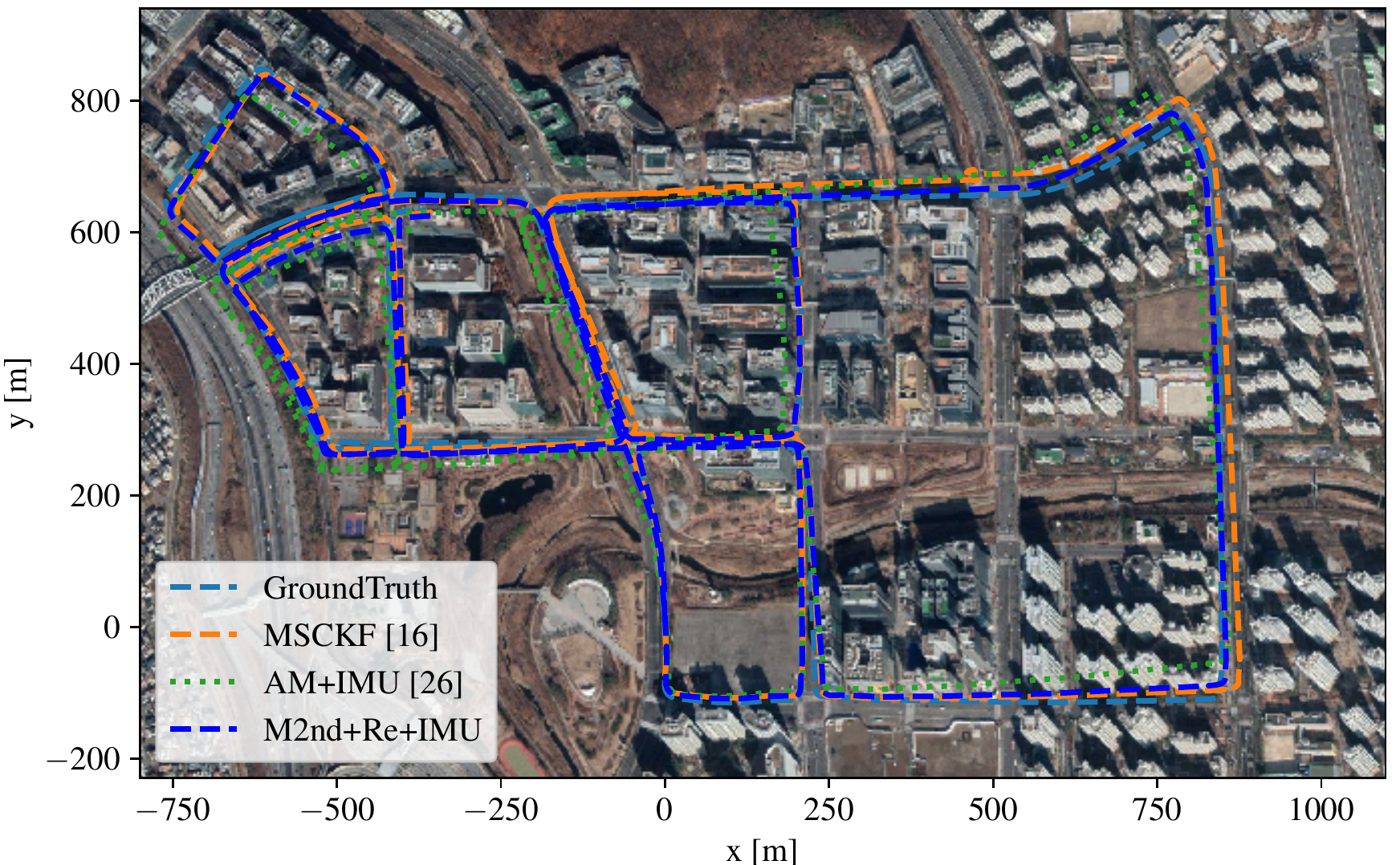} 
	}\hspace*{-0.0em}
	\caption{Ground truth and estimated trajectories of different approaches plotted on a Google map. The cyan dashed line corresponds to ground truth trajectory, orange dashed line to estimated one from MSCKF~\cite{li2013optimization}, green dotted one to AM+IMU\cite{zhang2019large}, and finally blue dashed one to the proposed method. The top figure shows the results on sequence\texttt{urban26-dongtan} and the bottom one on sequence \texttt{urban39-pankyo}.
	} \label{fig:kasit}
\end{figure}

\begin{figure}[htbp] 
	\subfigure{ 
		\includegraphics[width=0.48\linewidth]{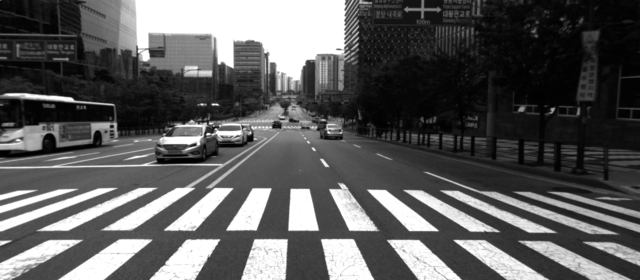} 
	}\hspace*{-0.8em}
	\subfigure{ 
		\includegraphics[width=0.48\linewidth]{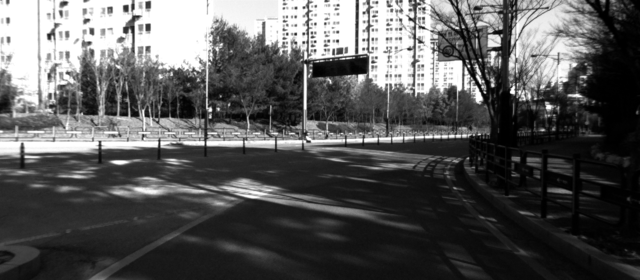} 
	}\vspace{-0.8em} 
	
	\subfigure{ 
		\includegraphics[width=0.48\linewidth]{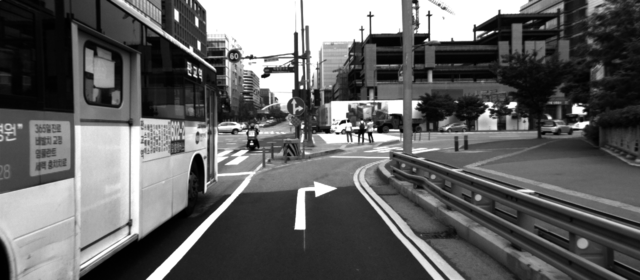} 
	}\hspace*{-0.8em}
	\subfigure{ 
		\includegraphics[width=0.48\linewidth]{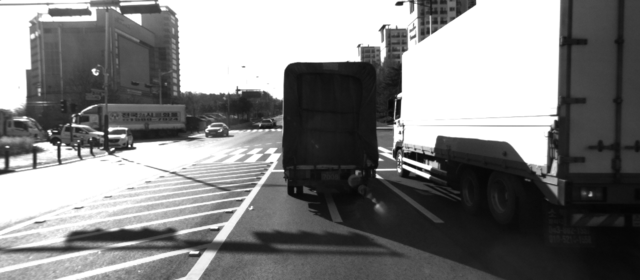} 
	}
	\caption{Representative images recorded from the left eye of the PointGrey camera in KAIST dataset~\cite{jeong2019complex}.}
	\label{fig:kaistfigure}
\end{figure}    

\definecolor{Gray1}{gray}{0.85}
\definecolor{Gray2}{gray}{0.65}
\newcolumntype{G}{>{\columncolor{Gray1}}c}
\newcolumntype{g}{>{\columncolor{Gray2}}c}
\renewcommand{\arraystretch}{1.20} 
\begin{table*}[!htbp] 
	\centering 
	\caption{Evaluations on highway scenarios: mean RMSE per 1-km for different algorithms in highway scenarios (\texttt{urban18}-\texttt{urban25} and
			\texttt{urban35}-\texttt{urban37}) of KAIST Urban dataset~\cite{jeong2019complex}.} 
	\label{table:kasit_highway} 
	\begin{tabular}{l 
			c G G G G G g g g g
		}  
		\toprule
		
		\multicolumn{1}{c}{{No. } }
		& \multicolumn{1}{c}{{Len. (km)} }
		& \multicolumn{1}{c} {\bf Mn }
		& \multicolumn{1}{c} {\bf M0th}
		& \multicolumn{1}{c} {\bf M1st}
		& \multicolumn{1}{c} {\bf M2nd}
		& \multicolumn{1}{c} {\bf M2nd+Re}
		& \multicolumn{1}{c} {\bf MSCKF\cite{li2013optimization}}
		& \multicolumn{1}{c} {\bf VINS-W\cite{wu2017vins}}
		& \multicolumn{1}{c} {\bf AM+IMU\cite{zhang2019large}}			
		& \multicolumn{1}{c} {\bf{M2nd+Re+IMU}}
		\\ 
		
		\midrule			
		\multicolumn{1}{c}{{w. IMU } }
		& \multicolumn{1}{c}{- }
		& \multicolumn{1}{c} { No}
		& \multicolumn{1}{c} { No}
		& \multicolumn{1}{c} { No}
		& \multicolumn{1}{c} { No}
		& \multicolumn{1}{c} { No}
		& \multicolumn{1}{c} { Yes}
		& \multicolumn{1}{c} { Yes}
		& \multicolumn{1}{c} { Yes}			
		& \multicolumn{1}{c} { Yes}
		\\ 
		
		\midrule
		
		{  \texttt{u18} }
		& 3.9
		&4.556  &7.727  &3.330  &3.372  &3.022  &6.544  &5.261  &3.328  & \bf{2.629 } \\

		\midrule
		{  \texttt{u19} } 
		& 3.0  
		&4.551  &5.903  &4.915  &4.543  &5.323  &8.298  &6.367  &7.375  & \bf{3.797 } \\

		\midrule
		{  \texttt{u20 } } 
		& 3.2
		& \bf{2.513 } &4.831  &3.031  &2.912  &2.746  &5.542  &4.824  &5.282  &2.717  \\

		\midrule
		{ \texttt{u21} } 
		& 3.7
		&3.444  &4.747  &3.791  &3.262  &4.255  &6.511  &6.014  &4.347  & \bf{2.787 } \\

		\midrule
		{  \texttt{u22} } 
		& 3.4
		&5.523  &5.208  &4.969  &4.425  &5.970  &6.934  &6.201  & \bf{2.626 } &4.221  \\

		\midrule
		{  \texttt{u23} } 
		& 3.4
		&6.285  &12.936  &4.607  &3.442  &3.823  &7.466  &5.452  &3.678  & \bf{2.684 } \\ 
		
		\midrule
		{  \texttt{u24} } 
		& 4.2
		&4.584  &7.569  &4.608  &4.109  &5.885  &6.011  &6.032  &5.568  & \bf{3.883 } \\

		\midrule
		{  \texttt{u25} } 
		& 2.5
		&7.695  &11.593  &6.546  &9.367  & \bf{1.870 } &6.394  &5.921  &2.019  &6.680  \\

		\midrule
		{  \texttt{u35} } 
		& 3.2
		&4.101  &5.536  &3.839  &3.417  &3.945  &8.332  &8.032  &3.826  & \bf{2.474 } \\

		\midrule
		{  \texttt{u36} } 
		& 9.0
		&6.354  &7.713  &4.663  &4.760  &5.500  &7.565  &7.608  &5.311  & \bf{3.887 } \\

		\midrule
		{  \texttt{u37} } 
		& 11.8
		&4.209  &2.999  &5.066  &3.054  &2.498  &5.255  &5.098  &2.685  & \bf{2.493 } \\

		\midrule
		{  mean} 
		& ---  
		&4.892  &6.978  &4.488  &4.242  &4.076  &6.805  &6.074  &4.186  & \bf{3.477 } \\ 
		
		\bottomrule  
	\end{tabular}
\end{table*}

\renewcommand{\arraystretch}{1.20} 
\begin{table*}[!htbp] 
	\centering 
	\caption{Evaluations on city scenarios: mean RMSE per 1-km for different algorithms in city scenarios (\texttt{urban26}-\texttt{urban34}, \texttt{urban38} and
			\texttt{urban39})
			of KAIST Urban dataset~\cite{jeong2019complex}.}
	\label{table:kasit_urban}   
	\begin{tabular}{l 
			c G G G G G g g g g
		}  
		\toprule
		
		\multicolumn{1}{c}{{No. } }
		& \multicolumn{1}{c}{{Len. (km)} }
		& \multicolumn{1}{c} {\bf Mn }
		& \multicolumn{1}{c} {\bf M0th}
		& \multicolumn{1}{c} {\bf M1st}
		& \multicolumn{1}{c} {\bf M2nd}
		& \multicolumn{1}{c} {\bf M2nd+Re}
		& \multicolumn{1}{c} {\bf MSCKF\cite{li2013optimization}}
		& \multicolumn{1}{c} {\bf VINS-W\cite{wu2017vins}}
		& \multicolumn{1}{c} {\bf AM+IMU\cite{zhang2019large}}			
		& \multicolumn{1}{c} {\bf{M2nd+Re+IMU}}
		\\ 
		
		\midrule			
		\multicolumn{1}{c}{{w. IMU } }
		& \multicolumn{1}{c}{- }
		& \multicolumn{1}{c} { No}
		& \multicolumn{1}{c} { No}
		& \multicolumn{1}{c} { No}
		& \multicolumn{1}{c} { No}
		& \multicolumn{1}{c} { No}
		& \multicolumn{1}{c} { Yes}
		& \multicolumn{1}{c} { Yes}
		& \multicolumn{1}{c} { Yes}			
		& \multicolumn{1}{c} { Yes}
		\\ 
		
		\midrule
		{  \texttt{u26} }
		& 4.0
		&3.426  &5.247  &3.692  &3.240  & 2.139 &4.129  &4.021  &3.742  &\bf{2.100}  \\

		\midrule
		{  \texttt{u27} } 
		& 5.4
		&4.285  &5.403  &4.767  &2.973  &2.449  &5.321  &4.913  &2.597  & \bf{2.439 } \\

		\midrule
		{  \texttt{u28} } 
		& 11.47 
		&4.428  &6.094  &5.892  &4.418  &3.227  &6.415  &6.021  &2.713  & \bf{2.352 } \\

		\midrule
		{ \texttt{u29} } 
		& 3.6
		&3.011  &3.421  &2.253  &2.814  &2.135  &3.714  &3.214  & \bf{1.989 } &2.057  \\

		\midrule
		{  \texttt{u30} } 
		& 6.0
		&3.920  &6.193  &5.023  &3.907  &2.954  &5.438  &4.982  &3.259  & \bf{2.442 } \\

		\midrule
		{  \texttt{u31} } 
		& 11.4
		&4.370  &6.264  &6.225  &4.374  &3.710  &7.002  &6.319  &3.890  & \bf{3.546 } \\

		\midrule
		{  \texttt{u32} } 
		& 7.1
		&3.364  &6.782  &5.775  &4.452  &2.288  &5.425  &4.923  &3.254  & \bf{2.137 } \\

		\midrule
		{  \texttt{u33} } 
		& 7.6  
		&3.312  &5.754  &4.140  &3.455  & \bf{2.461 } &5.837  &5.291  &3.533  &2.582 \\

		\midrule
		{  \texttt{u34} } 
		& 7.8 
		&5.139  &9.309  &7.514  &3.955  &3.664  &8.621  &8.237  &2.934  & \bf{2.422 } \\

		\midrule
		{  \texttt{u38} } 
		& 11.4
		&3.093  &7.869  &5.537  &3.213  & \bf{2.502 } &5.255  &5.032  &2.741  &2.519  \\

		\midrule
		{  \texttt{u39} } 
		& 11.1  
		&4.495  &7.068  &5.546  &4.270  &4.035  &5.111  &4.829  &2.420  & \bf{2.062 } \\

		\midrule
		{  mean} 
		& ---  
		&3.895  &6.309  &5.124  &3.734  &2.869  &5.661  &5.253  &3.007  & \bf{2.423 } \\

		\bottomrule  
	\end{tabular}
\end{table*}

\subsubsection{Large-scale Urban Tests for Ground Vehicles}
\label{sec:KAIST}
We first conducted real-world tests on ground vehicles in large-scale urban environments with KAIST dataset~\cite{jeong2019complex}.
Representative visualization of estimated trajectories on Google maps are shown in Fig.~\ref{fig:kasit}, and representative figures recorded in those datasets are shown in Fig.~\ref{fig:kaistfigure}.

Similar to the simulation tests, we implemented the same nine competing algorithms in this tests, including {\bf Mn }, {\bf M0th}, {\bf M1st}, {\bf M2nd}, {\bf M2nd+Re}, {\bf MSCKF}, {\bf VINS-W}, {\bf AM+IMU}, and {\bf M2nd+Re+IMU}.
The RMSEs of different algorithms are shown in Table~\ref{table:kasit_highway} and~\ref{table:kasit_urban} for highway and city scenarios in KAIST Urban dataset~\cite{jeong2019complex} respectively.
We note that different sequences have varied lengths which are listed in the second column of each table.
Additionally, since all competing algorithms are performing open-loop estimation, the computed errors are the mean of RMSEs over each 1-km segment.

This test demonstrates similar statistical results compared to the simulation tests.
First, from the first five light gray columns in Table~\ref{table:kasit_highway} and~\ref{table:kasit_urban},  by using `poor' parameterization for the motion manifold, the estimation accuracy will be limited. In fact, when parameterized by zeroth order ({\bf M0th}) or first order ({\bf M1st}) polynomials, using the motion manifold for estimation will be worse than not using it ({\bf Mn}), especially in the city scenarios. 
This is due to the fact that the motion manifold is typically \rev{complex} outdoors and the simple representations are not enough to model the manifold properly, which will introduce invalid constraints and deteriorate the estimator.
However, when proper modeling (i.e. the proposed method ({\bf M2nd})) is used, better estimation precision can be achieved, which validates the most important assumption in our paper: with properly designed motion manifold representation, the estimation accuracy can be improved.
In addition, by comparing the estimation errors of {\bf M2nd} and {\bf M2nd+Re} in all sequences, we noticed that when the manifold re-parameterization is not used, the errors increase significantly. This validates our theoretical analysis in Sec.~\ref{sec:re-param} that the re-parameterization is a necessary step to ensure manifold propagation errors can be characterized statistically.
Moreover, let us review the data in four dark gray columns, corresponding to the competing methods when an IMU sensor is used. In fact, pose estimates from {\bf MSCKF\cite{li2013optimization}}, {\bf VINS-W~\cite{wu2017vins}} and {\bf AM+IMU~\cite{zhang2019large}} are consistently worse than the proposed methods.
All these facts validate our theoretical analysis that by properly modeling and using the motion manifold, the estimation errors can be clearly reduced. 
\rev{We note that, there are also failure cases for the proposed manifold representation, such as using it for robots on an uneven farmland road. However, for urban applications with well-constructed roads, our method can be widely used.}

\begin{figure*}[ht] 
	\begin{minipage}{.35\linewidth}
		\hspace*{-0.1em}
		\subfigure{
			\includegraphics[width=2.5in]
			{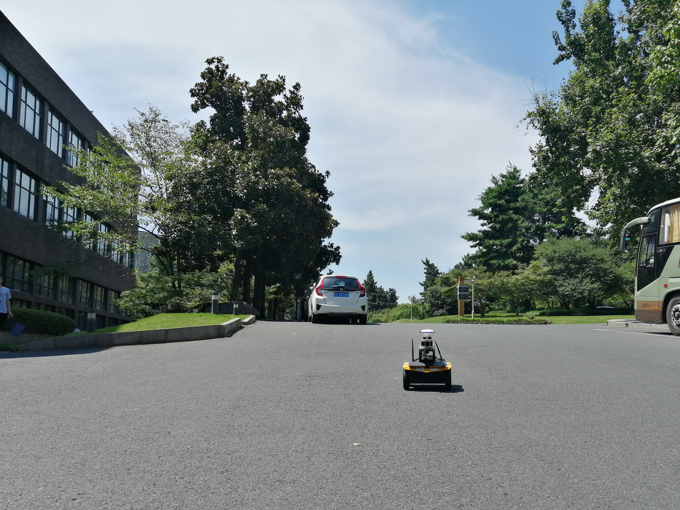}
		}
	\end{minipage}
	\begin{minipage}{.45\linewidth}
		\subfigure{ 
			\includegraphics[width=1.5in]{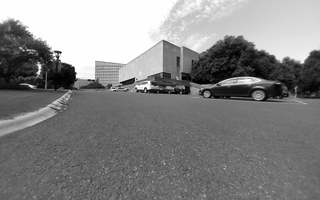} 
		}\hspace*{-0.8em}
		\subfigure{ 
			\includegraphics[width=1.5in]{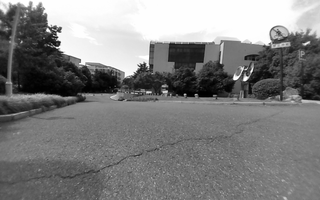} 
		}\hspace*{-0.8em}
		\subfigure{ 
			\includegraphics[width=1.5in]{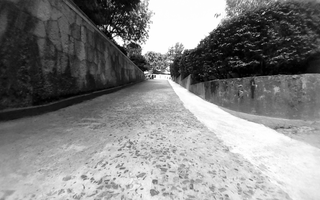} 
		}\hspace*{-0.8em}
		\vspace{-0.8em} 
		
		\subfigure{ 
			\includegraphics[width=1.5in]{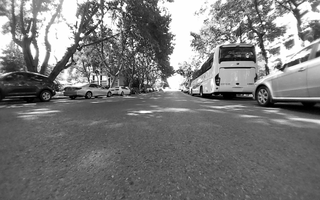} 
		}\hspace*{-0.8em}
		\subfigure{ 
			\includegraphics[width=1.5in]{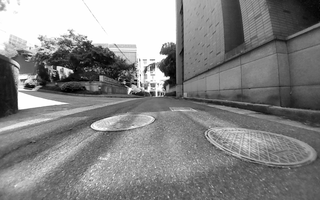} 
		}\hspace*{-0.8em}
		\subfigure{ 
			\includegraphics[width=1.5in]{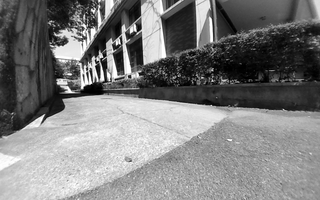} 
		}\hspace*{-0.8em}
	\end{minipage}
	\caption{Representative visualization of outdoor experiment scenario.
	} \label{fig:zju}
\end{figure*}

\begin{figure}[tbp]
	\subfigure{
		\includegraphics[width=\columnwidth]
		{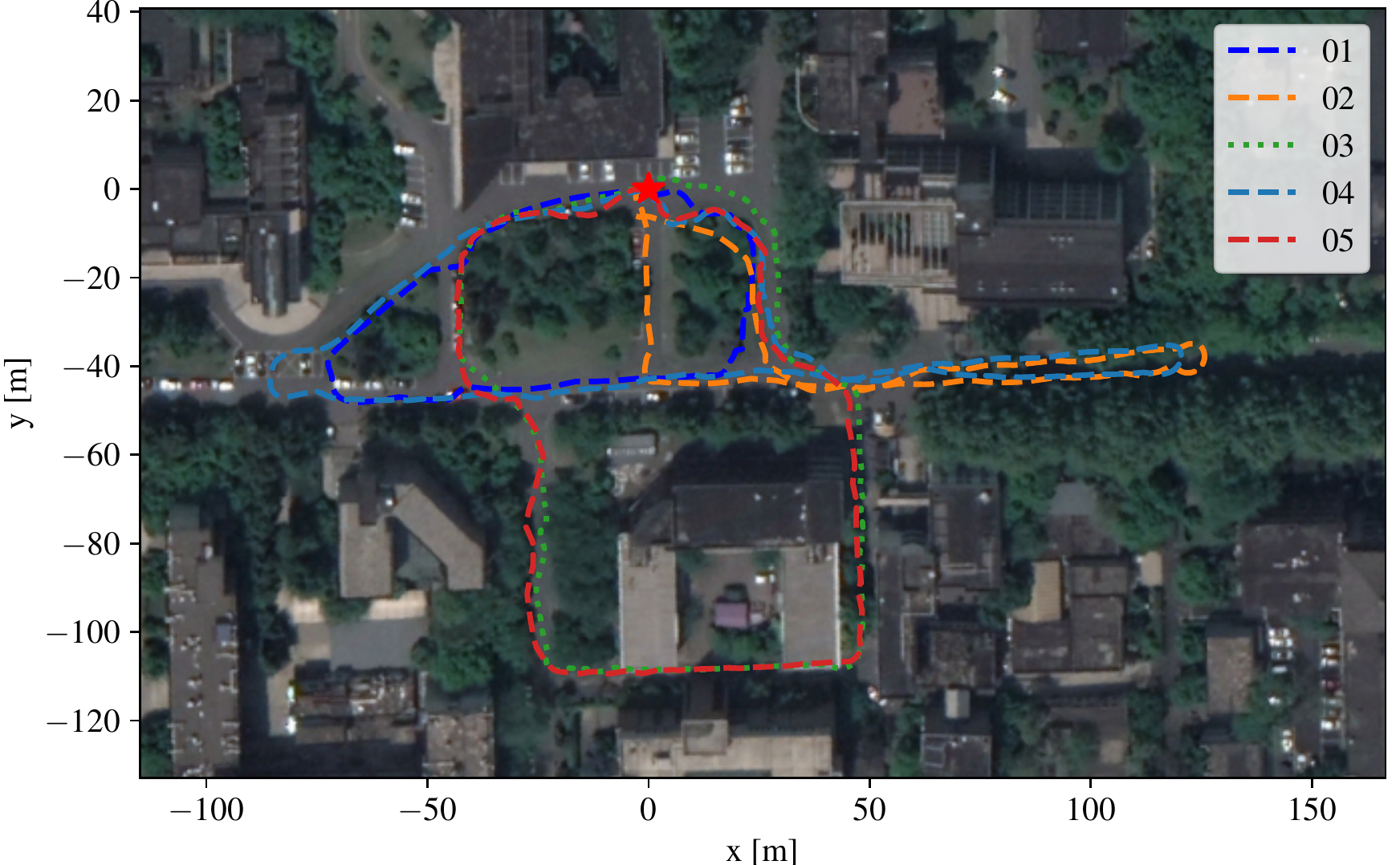}
	}
	\caption{\rev{The trajectories of the five sequences (01 to 05) with different colors overlaid on the map of ZJU campus. The identical starting position for all sequences is marked by a red star.}}\label{fig:plot_zju}
\end{figure}

\begin{table*}[ht]  
	\centering 
	\fontsize{8.0}{8.0}\selectfont  
	\begin{threeparttable}  
		\caption{Final positional drift for competing algorithms in Campus dataset.}  
		
		\label{table:pri sensors}
		\begin{tabular}{l 
				c c c c c c c
			}  
			\toprule
			\multicolumn{1}{c}{\multirow{2}*{No.} }
			& \multicolumn{1}{c}{\multirow{2}*{Length(m)} }
			& \multicolumn{5}{c}{Final Positional Drift (m)} \\
			\cline{3-8} 
			\multicolumn{1}{c}{}
			&{}
			& \multicolumn{1}{c} {\bf{M2nd+Re+IMU}}
			& \multicolumn{1}{c} {\bf{M2nd+Re}}
			& \multicolumn{1}{c} {\bf{M2nd}}
			& \multicolumn{1}{c} {\bf AM+IMU\cite{zhang2019large}}
			& \multicolumn{1}{c} {\bf VINS-W\cite{wu2017vins}}
			& \multicolumn{1}{c} {\bf MSCKF\cite{li2013optimization}}
			\\ 
			
			\midrule
			{  \bf{ 01 } } 
			& 246.2
			& 1.7804 & 1.9006 &  2.2940 & \bf{1.1568} &1.8643 &1.9558 \\
			
			\midrule
			{  \bf{ 02 } } 
			& 332.4
			& \bf{3.3028} & 5.7115 & 6.6152 & 3.8500 & 5.2419 &5.4742 \\
			
			\midrule
			{  \bf{ 03 } } 
			& 359.8
			& \bf{2.0334}&  2.8503  & 11.5017 & 2.7974 & 3.3949 & 3.4796\\
			
			\midrule
			{ \bf{ 04 } } 
			& 472.5
			& \bf{2.4266} & 3.6845 & 7.6708 & 2.9968 & 4.2351 & 4.2742\\
			
			\midrule
			{  \bf{ 05 } } 
			& 364.6
			& \bf{1.8863} & 3.5367 & 47.9063 & 2.7326 &5.5932 &5.8041\\
			
			\bottomrule  
		\end{tabular}
	\end{threeparttable}  
\end{table*}

\subsubsection{Ground Robot Tests in a University Campus}
\label{sec:exp2} 

The next experiment is to evaluate the overall pose estimation performance of the proposed method, 
	compared to computing algorithms similar to previous tests, i.e., {\bf M2nd+Re}, {\bf M2nd}, {\bf MSCKF}, {\bf VINS-W}, {\bf AM+IMU}, and {\bf M2nd+Re+IMU}. 
	Since methods of {\bf Mn}, {\bf M0th}, {\bf M1st} in previous tests are mainly used as `ablation study' instead of demonstrating our core contributions, those methods are omitted in this test to allow readers to focus the `real' competing methods.
Evaluations were conducted on five sequences, with selective images and trajectories are shown in Fig.~\ref{fig:zju} and Fig.~\ref{fig:plot_zju}.
The final positional drift errors were computed as the metric for different methods. 

Table~\ref{table:pri sensors} shows the final positional drift for all methods  and over all testing sequences. The proposed method clearly achieves the best performance.
In addition, similar to our simulation tests, 
those results show that when an IMU is used as an additional sensor for estimation, the performance of the proposed algorithm can be further improved.
We also note that, the re-parametrization module introduced in Sec.~\ref{sec:re-param} is critical to guarantee accurate estimation performance for the system. 

\subsection{\rev{Runtime} Analysis}
In addition to the accuracy evaluation, we also try to analyze the computational cost of the proposed algorithmic modules. We will show that by integrating our modules, the computational cost of sliding-window estimator remains almost unchanged.
The tests in this section were performed on an Intel Core i7-4790 CPU running at 2.90GHz. Our code was written in C++, and the sliding-window size of our implementation was set to be $8$.

Specifically, we evaluated the run time of each critical sub-modules in four algorithms: {\bf Mn}, {\bf M2nd+Re}, {\bf M2nd+Re+IMU}, and {\bf IMU Method}. The last method is to use the proposed sliding-window algorithm and implementation, but with IMU and monocular camera measurements only. Table~\ref{table:time} shows results for four main sub-modules in a single sliding window update cycle: feature extraction, feature tracking, iterative optimization, and probabilistic marginalization (in the proposed method, the re-parameterization time is counted as part of the marginalization process). In Table~\ref{table:time}, the reported values for each module are computed by averaging results from $2500$ sliding-window update cycles.
Those results demonstrate that the \rev{runtime} for all four methods are similar. In fact, the computational costs of sliding-window optimization are primarily determined by 1) image processing time, 2) size of the sliding window, and 3) number and track length of processed features~\cite{li2013optimization}.
By using our method, the \rev{increased} computational complexity is \rev{almost negligible}. 

\begin{table}[t]  
	\centering 
	\begin{threeparttable}  
		\caption{\rev{Runtime statistics, i.e., mean and (standard deviation), of the main sub-modules.}}  
		\label{table:time}
		\setlength{\tabcolsep}{5pt}
		\begin{tabular}{c c c c c
			}  
			\toprule
			\multicolumn{1}{c}{\multirow{2}*{\bf Modules} }
			& \multicolumn{4}{c}{Computation Time (ms)} \\
			\cline{2-5} 
			\multicolumn{1}{c}{}
			& \multicolumn{1}{c} {\bf{ Mn}}
			& \multicolumn{1}{c} {\bf  {M2nd+Re}}
			& \multicolumn{1}{c} {\bf  {M2nd+Re+IMU}}
			& \multicolumn{1}{c} {\bf  {IMU M.}} \\ 
			
			\midrule
			{  { Feat. Detect.} } 
			&9.02(1.18) & 9.14(1.32) & 9.15(1.26) &9.09(1.11) \\

			\midrule
			{  { Feat. Track. } } 
			&3.64(0.72) & 3.78(0.69) & 3.73(0.67) & 3.77(0.72) \\

			\midrule
			{  { Opt. } } 
			& 7.95(2.63) & 9.12(2.99) & 10.01(3.04) & 9.77(3.01) \\
			
			\midrule
			{  { Marg. } } 
			& 1.70(0.49) & 1.82(0.48) & 2.38(0.49)  & 2.18(0.46) \\
			
			\bottomrule  
		\end{tabular}
	\end{threeparttable}  
\end{table}

\section{CONCLUSIONS}
In this paper, we proposed \rev{an accurate} pose estimation algorithm dedicatedly designed \rev{for non-holomonic} ground robots \rev{with measurements from the wheel encoder and an exteroceptive sensor}. Since ground robots typically navigate on a local motion manifold, it is possible to exploit this fact to derive kinematic constraints and use proper low-cost sensors in estimation algorithms.
Specifically, we explicitly modelled the motion manifold by continuously differentiable parametric representation. Subsequently, we proposed a method for performing full 6D pose integration using both wheel odometer measurements and the motion manifold in closed-form. We also analyzed the estimation errors caused by the manifold representation and performed re-parameterization periodically to achieve error reduction.
Finally, \rev{we validate the proposed method in} 
an optimization-based sliding window estimator for fusing measurements from a monocular camera, wheel odometer, and optionally an IMU.
By extensive experiments conducted from both simulated and real-world data, we show that the proposed method outperforms competing state-of-the-art algorithms by a significant margin \rev{in pose estimation accuracy}.

\begin{appendices}
\section{Derivative of Orientation Error State}\label{sec:orentation error}
In this section, we describe the detailed derivation of the orientation error state differential equation, i.e., Eq.~\eqref{eq:derivative theta}.
To start with, we substitute Eq.~\eqref{eq:rot error} into Eq.~\eqref{eq:rot dir}, which leads to:
\begin{align}
\label{eq:rot direst}
&\frac{\partial \bigg{(}
	{^{\bG}_{\bO} \hat{\bR}} 
	\left( 
	{\mathbf I + \lfloor \delta \btheta
		\rfloor}
	\right)\bigg{)}}{\partial t} = 
{^{\bG}_{\bO} \hat{\bR}}  \left( 
{\mathbf I + \lfloor \delta \btheta
	\rfloor}
\right)
\lfloor {^{\bO} \bomega}  \rfloor \\
\Rightarrow& 
{^{\bG}_{\bO} \hat{{\bR}}} 
\lfloor {^{\bO} \hat{\bomega}}  \rfloor
\! \left( 
{\mathbf I \!+\! \lfloor \delta \btheta
	\rfloor}
\right)\!+\!
{^{\bG}_{\bO} \hat{\bR}} 
\lfloor \dot{\delta \btheta}
\rfloor\!
= \!
{^{\bG}_{\bO} \hat{\bR}}  \left( 
{\mathbf I \!+\! \lfloor \delta \btheta
	\rfloor}
\right)\!
\lfloor {^{\bO} \hat{\bomega} \!+\!\! {}
	^{\bO}\tilde{\bomega}}  \rfloor \\
\Rightarrow& 
\lfloor {^{\bO} \hat{\bomega}}  \rfloor
\! \left( 
{\mathbf I \!+\! \lfloor \delta \btheta
	\rfloor}
\right)\!+\! 
\lfloor \dot{\delta \btheta}
\rfloor\!
= \!
\left( 
{\mathbf I \!+\! \lfloor \delta \btheta
	\rfloor}
\right)\!
\lfloor {^{\bO} \hat{\bomega} \!+\!\! {}
	^{\bO}\tilde{\bomega}}  \rfloor
\end{align}
By applying the quadratic error approximation $\lfloor \delta \btheta
\rfloor \lfloor {}
^{\bO}\tilde{\bomega}
\rfloor \simeq \mathbf 0$, the above equation can be written as:
\begin{align}
\lfloor \dot{\delta \btheta}
\rfloor\!
&\simeq \!
\left( 
{\mathbf I \!+\! \lfloor \delta \btheta
	\rfloor}
\right)\!
\lfloor {^{\bO} \hat{\bomega} \!+\!\! {}
	^{\bO}\tilde{\bomega}}  \rfloor -
\lfloor {^{\bO} \hat{\bomega}}  \rfloor
\! \left( 
{\mathbf I \!+\! \lfloor \delta \btheta
	\rfloor}
\right) \\
&\simeq
\lfloor \delta \btheta
\rfloor \lfloor {^{\bO} \hat{\bomega}}  \rfloor - 
\lfloor {^{\bO} \hat{\bomega}}  \rfloor
\lfloor \delta \btheta
\rfloor	+ 	
\lfloor {^{\bO}\tilde{\bomega}}  \rfloor
\label{eq:aaa}
\end{align}
Finally, by using the property of skew-symmetric matrix:
\begin{align}
\lfloor \mathbf a
\rfloor \lfloor \mathbf b  \rfloor - 
\lfloor \mathbf b \rfloor
\lfloor \mathbf a
\rfloor = 
\mathbf b \mathbf a^T - {}
\mathbf a \mathbf b^T = 
\lfloor \lfloor \mathbf a  \rfloor \mathbf b
\rfloor
=
-
\lfloor \lfloor \mathbf b  \rfloor \mathbf a
\rfloor
\end{align}
Eq.~\eqref{eq:aaa} becomes:
\begin{align}
\dot{\delta \btheta} = -
\lfloor {^{\bO} \hat{\bomega}}  \rfloor \delta \btheta +
{^{\bO}\tilde{\bomega}}
\end{align}
and 
\begin{align}
{^{\bO}\tilde{\bomega}} = 
\mathbf J_\bst \bstt
+
\mathbf J_n \mathbf{n}_{\omega}
\end{align}
This completes the derivation.

\section{Jacobian of the Inferred Angular Velocity}\label{sec:jacobian omega}
In this section, we describe the detailed derivation of Jacobians in Eq.~\eqref{eq:derivative theta}.
Recall the inferred local angular velocity in Eq.~\eqref{eq: final omega}:
\begin{align}
\label{eq: omega error11}
^{\bO(t)}
\hat{\bomega} = 
\begin{bmatrix}
\frac{1}{\left\| \nabla \hat{\mathcal{M}}(t) \right\|}
\mathbf e^T_{12}
\big{\lfloor}
{\mathbf e_3}
\big{\rfloor}
\big{(} {^{\mathbf G}_{\mathbf O(t)} 
	\hat{\bR}^T} \cdot \hat{\dot{\nabla \mathcal{M}}} (t)
\big{)} \\
\omega_o (t)
\end{bmatrix}
\end{align}
where $\nabla \hat{\mathcal{M}}(t)$ and $\hat{\dot{\nabla \mathcal{M}}} (t)$ represent the gradient and it time derivative of the equality manifold constraint evaluated at current state estimate.
Based on Eq.~\eqref{eq:manifold}, we can derive:
\begin{align}
\nabla \mathcal{M} = 
\begin{bmatrix}
\mathbf A 
\begin{bmatrix}
{}^\bG \bp _{\bO_x} \\ {}^\bG \bp _{\bO_y} 
\end{bmatrix}
+ 
\mathbf B \\
1
\end{bmatrix},\,\,
\dot{\nabla \mathcal{M}}
= 
\begin{bmatrix}
\mathbf A 
\begin{bmatrix}
^\bG v_x \\ ^\bG v_y
\end{bmatrix}
\\
0
\end{bmatrix}
\end{align}
or equivalently:
\begin{align}
\nabla \mathcal{M} \! \!  =\!  \!  
\bar{\mathbf A}
{^\bG \bp _\bO} \!  \!+\!\!  
\bar{\mathbf B}\!\!  +\! \!  \mathbf e_3,
\dot{\nabla \mathcal{M}} \! \!  =\!  \!  
\bar{\mathbf A}
{^\bG \bv _\bO},
\bar{\mathbf A} \! \!  =\!  \!  
\begin{bmatrix}
\bA & \mathbf 0 \\
\mathbf 0^T & 0
\end{bmatrix},
\bar{\mathbf B} \! \!  = \! \!  
\begin{bmatrix}
\bB \\ 0
\end{bmatrix}
\end{align}
By denoting $\bE = 
\mathbf e^T_{12}
\big{\lfloor}
{\mathbf e_3}
\big{\rfloor}$, 
${\bD} = 
{_{\mathbf G}^{\mathbf O} 
	{\bR}} \cdot \bar{\bA}
\cdot
{_{\mathbf G}^{\mathbf O} 
	{\bR}^T}$ and dropping $(t)$ for simplicity, we write Eq.~\eqref{eq: final omega} as:
\begin{align}
\label{eq: new final omega}
^{\bO}\bomega = 
\begin{bmatrix}
\frac{1}{\left\| \nabla \mathcal{M} \right\|}
\bE \bD \left( ^{\bO}\bv - {} \bn_v\right) \\
\omega_o - 
n_{\omega o}
\end{bmatrix}
\end{align}
To compute the state-transition matrix, we first note that:
\begin{align}
\label{eq:dev rule}
&\frac{
	\partial ||\mathbf y||}
{
	\partial \mathbf y} = 
\frac{
	\mathbf y^\top}
{||\mathbf y||},\,\,\,\,
\frac{
	\partial \frac{1}{||\mathbf y||}}
{
	\partial \mathbf y} = 
-\frac{
	\mathbf y^\top}
{||\mathbf y||^3},\,\,\,\, \text{and} \notag \\
&\frac{
	\partial
	\frac{\mathbf k}{||\mathbf y||}
}{
\partial \mathbf y} = 
-\frac{
	\mathbf k \mathbf y^T}
{||\mathbf y||^3},\,\,\,\,
\frac{
	\partial
	\frac{\mathbf k}{||\mathbf y||}
}{
\partial \mathbf x} = 
-
\frac{
	\mathbf k \mathbf y^T}
{||\mathbf y||^3}
{\frac{\partial \mathbf y}{\partial \mathbf x}}
\end{align}
By employing Eq.~\eqref{eq:dev rule}, we have:
\begin{align}
\frac{\partial {^{\bO}\bomega}}{\partial {^\bG \bp _\bO}} 
= 
\begin{bmatrix}
-\bE \hat{\bD} {}{\mathbf e_1 v_o}  \frac{\nabla\hat{\mathcal{M}}^\top}{\left\| \nabla\hat{\mathcal{M}} \right\|^3} \bar{\bA} \\
\mathbf{0}
\end{bmatrix}
\end{align}
and
\begin{align}
\frac{\partial {^{\bO}\bomega}}{\partial \delta \btheta} 
= 
\begin{bmatrix}
\frac{\bE}{\left\| \nabla\hat{\mathcal{M}} \right\|}\cdot \left( \lfloor \hat{\bD} {}{\mathbf e_1 v_o} \rfloor - \hat{\bD} \lfloor {}{\mathbf e_1 v_o} \rfloor \right)\\
\mathbf{0}
\end{bmatrix}
\end{align}
finally
\begin{align}
\frac{\partial {^{\bO}\bomega}}{\partial  \bm} 
=
\begin{bmatrix}
-\bE \hat{\bD} {}{\mathbf e_1 v_o}  \frac{\nabla\hat{\mathcal{M}}^\top}{\left\| \nabla\hat{\mathcal{M}} \right\|^3} \hat{\boldsymbol{\Gamma}} + \frac{\bE}{\left\| \nabla\hat{\mathcal{M}} \right\|} {^{\mathbf O}_{\mathbf G} 
	\hat{\bR}} \hat{\boldsymbol{\Upsilon}}\\
\mathbf{0}
\end{bmatrix}
\end{align}
with
\begin{align}
\boldsymbol{\Gamma} = 
\begin{bmatrix}
0& 1& 0& {}^\bG \bp _{\bO_x} & {}^\bG \bp _{\bO_y}& 0  \\
0& 0& 1& 0& {}^\bG \bp _{\bO_x} & {}^\bG \bp _{\bO_y} \\
0& 0& 0& 0& 0& 0
\end{bmatrix}
\end{align}
and
\begin{align}
\boldsymbol{\Upsilon} = 
\begin{bmatrix}
0& 0& 0& {}^\bG \bv _{\bO_x} & {}^\bG \bv _{\bO_y}& 0 \\
0& 0& 0 & 0& {}^\bG \bv _{\bO_x} & {}^\bG \bv _{\bO_y} \\
0& 0& 0& 0& 0& 0
\end{bmatrix}
\end{align}
To summarize the Jacobians with respect to the state, we have:
\begin{align}
\mathbf{J}_\bst = 
\begin{bmatrix}
\frac{\partial {^{\bO}\bomega}}{\partial {^\bG \bp _\bO}} & \frac{\partial {^{\bO}\bomega}}{\partial \delta \btheta} & \frac{\partial {^{\bO}\bomega}}{\partial  \bm}    
\end{bmatrix}
\end{align}
On the other hand, $\mathbf{J}_n$ can be easily computed by taking derivative with respect to $\bn = \begin{bmatrix}
n_{vo} & n_{\omega o}
\end{bmatrix}^T$.
\begin{align}
\mathbf{J}_n = 
-\begin{bmatrix}
\frac{1}{\left\| \nabla \hat{\mathcal{M}} \right\|}
\bE \hat\bD \mathbf e_1 & \mathbf{0}_{2 \times 1}\\
0 & 1 & 
\end{bmatrix}
\end{align}

\section{Derivation on Manifold Re-Parameterization}
\label{app:reparam-app}
In this section, we provide the detailed derivations of Eq.~\eqref{eq:re-param}. To simplify our analysis, we here use $\mathbf m_o = \bm (t_k)$ and 
$\mathbf m_1 = \bm (t_{k+1})$ for abbreviation.
In our work, we define the local manifold representation at $t_k$ and $t_{k+1}$ as:
\begin{align}
\label{eq:manifold new xo}
&z \!+\!
c \!+\! \mathbf B^T 
{\Delta{\mathbf p}_o}
\!+\!\frac{1}{2}
{\Delta{\mathbf p}^T_o}
\mathbf A 
{\Delta{\mathbf p}_o} \!=\! 0,\,{\Delta{\mathbf p}_o} \!=\! 
\bR _o\left( \begin{bmatrix}
x \\ y
\end{bmatrix}
\!-\! 
\begin{bmatrix}
x_o \\ y_o
\end{bmatrix} \right) \\
\label{eq:manifold new x1}
&z \!+\!
c \!+\! \mathbf B^T 
{\Delta{\mathbf p}_1}
\!+\! \frac{1}{2}
{\Delta{\mathbf p}^T_1}
\mathbf A 
{\Delta{\mathbf p}_1}  \!=\! 0,\,{\Delta{\mathbf p}_1} \!=\! 
\bR _1
\left( \begin{bmatrix}
x \\ y
\end{bmatrix}
\!-\! 
\begin{bmatrix}
x_1 \\ y_1
\end{bmatrix} \right)
\end{align}
where $\bR_o, \bR_1 \in \mathbb{R}^{2 \times 2}$ and 
the detailed formulation of $\mathbf A$ and 
$\mathbf B$ can be found in
Eq.~\eqref{eq:man para0} and Eq.~\eqref{eq:man para}. We further define  $\bp^T_o = \begin{bmatrix}
	x_o & y_o
	\end{bmatrix}$ and $\bp^T_1 = \begin{bmatrix}
	x_1 & y_1
	\end{bmatrix}$ with a slight abuse of notation.
We note that, Eqs.~\eqref{eq:manifold new xo} -~\eqref{eq:manifold new x1} should hold 
for all $\mathbf p$, and thus the corresponding coefficients for $\mathbf p$ should be the same in both equations. 
Considering the equality on the second order coefficients leads to:
\begin{align}
&\bR_o^T \mathbf A \bR_o = 
\bR_1^T \mathbf A_{new} \bR_1 \\
\Rightarrow&
\mathbf A_{new} = 
\dbR^T \mathbf A \dbR,\,\,\, \text{with} \,\,\,
\dbR = \bR_o \bR_1^{-1}
\end{align}
and the first order ones:
\begin{align}
&\mathbf B^T \bR_o - 
\bp_o^T \bR^T_o \mathbf A \bR_o
=
\mathbf B_{new}^T \bR_1 - 
\bp_1^T \bR^T_1 \mathbf A_{new} \bR_1 \\
\Rightarrow&
\mathbf B_{new}^T \!=\! 
\mathbf B^T \dbR - 
\bp_o^T  \bR_o^T \mathbf A \dbR + 
\bp_1^T  \bR_o^T \mathbf A \dbR \\
\Rightarrow&
\mathbf B_{new} \!=\! 
\dbR^T \mathbf B  \!+ \!
\dbR^T \mathbf A^T  \dbp, \,
\text{with}\, \dbp \!=\! \bR_o \left(
\bp _1 \!-\! \bp _o \right)
\end{align}
finally the constant value:
\begin{align}
&c - \mathbf B^T \bR_o \bp _o + \frac{1}{2}
\bp^T _o \bR_o^T \mathbf A \bR_o \bp _o \notag \\
&\qquad = c_{new} - \mathbf B_{new}^T \bR_1 \bp _1 + \frac{1}{2}
\bp^T _1 \bR_1^T \mathbf A_{new}\bR_1  \bp _1 \\
\Rightarrow& 
c_{new} \!=\! c \!\!+\!  \!
\mathbf B^T \dbR \bR_1 \bp_1 \!\!+\!\! 
(\bp _1 \!\!-\!\!  \bp _o)^T \bR_o^T \mathbf A \dbR \bR_1 \bp_1 \!\!-\!\!
\mathbf B^T \bR_o \bp _o \notag \\ 
&\qquad - \frac{1}{2}
\bp^T _1 \bR_o^T \mathbf A \bR_o  \bp _1 + 
\frac{1}{2}
\bp^T _o \bR_o^T \mathbf A \bR_o  \bp _o \\
\Rightarrow& 
c_{new} = c + \mathbf B^T  \dbp +
\frac{1}{2}
\dbp^T  \mathbf A \dbp
\end{align}
In matrix form representation, we have:
\begin{align}
\label{eq:re-param x}
\mathbf m_{new} = 
\underbrace{\begin{bmatrix}
	1 & \dbp^T & \boldsymbol{\gamma}^T \\
	\mathbf 0 & \dbR^T & \boldsymbol{\Xi} \\
	\mathbf 0 & \mathbf 0 & 
	\boldsymbol{\Psi}
	\end{bmatrix}}_{\boldsymbol{\Lambda}_1}
\mathbf m
\end{align}
with
\begin{align}
\boldsymbol{\gamma}^T = \
\begin{bmatrix}
0.5 \delta x ^ 2 &
\delta x \delta y&
0.5 \delta y ^ 2 
\end{bmatrix},\,
\boldsymbol{\Xi} = 
\dbR^T 
\begin{bmatrix}
\delta x & \delta y & 0 \\
0 & \delta x & \delta y 
\end{bmatrix}
\end{align}
and 
\begin{align}
\boldsymbol{\Psi}\! = \!\begin{bmatrix}
\dr ^2_1 & 2 \dr_1 \dr_3 & \dr_3^2 \\
\dr _1 \dr _2 & \dr_1 \dr_4 + \dr_2 \dr_3 & \dr _3\dr _4  \\
\dr ^2_2 & 2 \dr_2 \dr_4 & \dr_4^2 
\end{bmatrix},
\dbR \!=\! \begin{bmatrix}
\dr_1 & \dr_2 \\
\dr_3 & \dr_4
\end{bmatrix}
\end{align}
This completes our derivation.

We next discuss the choices of parameters in $\bp _o$, $\bp _1$, $\bR_o$, and $\bR_1$. In this work, we choose the first estimate of $^\bG \bp _{\bO(t_k)}$ for $\bp _o$ and 
$^\bG \bp _{\bO(t_{k+1})}$ for $\bp _1$. 
Based on our theoretical analysis in Sec.~\ref{sec:re-param} and experimental results,
we show that our parameter choices of $\bp _o$ and $\bp _1$ are \rev{meaningful}, resulting in significant performance gains.
However, for $\bR_o$ and $\bR_1$, we tried two sets of choices, by extracting the corresponding elements in {\em yaw} of the estimate of $^\bG _{\bO(t_k)} \bR^T $ and $^\bG _{\bO(t_{k+1})} \bR^T $ and by simply setting $\mathbf I_2$ matrices.
The results show that the two choices have similar performances. Thus for the experimental results reported in this paper, we use $\mathbf I_2$ matrices as the orientation parameters for the simplicity.
\end{appendices}
	
\bibliographystyle{unsrt}%
\bibliography{ref}

\end{document}